\documentclass[10pt,twocolumn,letterpaper]{article}

\usepackage{cvpr}
\usepackage{times}
\usepackage{epsfig}
\usepackage{graphicx}
\usepackage{amsmath}
\usepackage{amssymb}

\usepackage{kotex} 
\newcommand{\RNum}[1]{\uppercase\expandafter{\romannumeral #1\relax}}
\usepackage{subfig}
\usepackage{stackengine}
\usepackage{booktabs}
\usepackage{multirow}
\usepackage[breaklinks=true,bookmarks=false]{hyperref}

\cvprfinalcopy 

\def\cvprPaperID{****} 

\begin{document}

\title{5D Light Field Synthesis from a Monocular Video}

\author{Kyuho Bae\textsuperscript{1},
Andre Ivan\textsuperscript{1},
Hajime Nagahara\textsuperscript{2},
and In Kyu Park\textsuperscript{1}\\
{\tt\small \{kyuho1104@gmail.com, andreivan13@gmail.com, nagahara@ids.osaka-u.ac.jp, pik@inha.ac.kr\}}
\\
\textsuperscript{1}Dept. of Information and Communication Eng., Inha University, Incheon 22212, Korea\\
\textsuperscript{2}Institute for Datability Science, Osaka University, Osaka, Japan
}
\maketitle

\begin{abstract}
   Commercially available light field cameras have difficulty in capturing 5D (4D + time) light field videos. They can only capture still light filed images or are excessively expensive for normal users to capture the light field video.
   To tackle this problem, we propose a deep learning-based method for synthesizing a light field video from a monocular video.
   We propose a new synthetic light field video dataset that renders photorealistic scenes using UnrealCV rendering engine because no light field dataset is avaliable.
   The proposed deep learning framework synthesizes the light field video with a full set (9$\times$9) of sub-aperture images from a normal monocular video.
   The proposed network consists of three sub-networks, namely, feature extraction, 5D light field video synthesis, and temporal consistency refinement.
   Experimental results show that our model can successfully synthesize the light field video for synthetic and actual scenes and outperforms the previous frame-by-frame methods quantitatively and qualitatively.
   The synthesized light field can be used for conventional light field applications, namely, depth estimation, viewpoint change, and refocusing.
\end{abstract}

\section{Introduction}
The recent decade witnessed a rapid growth of light field technology which has received substantial interest in the fields of computer vision and graphics.
Different from a conventional image, a light field image captures 4D light information of directional rays through the main lens of the camera instead of accumulating them.
Direct applications, depth image estimation~\cite{schilling2018trust,shin2018epinet,williem2016robust,park2017robust}, image refocusing~\cite{ng2005light}, saliency detection~\cite{li2014saliency}, and view-point change are performed using only a single shot of light field image as a post-capturing process.
Traditionally, light field images are captured using a plenoptic camera consisting of a microlens array or a multicamera array~\cite{wilburn2005high}.
However, Lytro's camera, which was commercially available for general users, is no longer avaliable in the market.
The only available camera is Raytrix~\cite{Raytrix_2013}, which can capture light field videos.
However, it is used for industrial and research purposes. Therefore, it is much expensive for general users for daily use.
To overcome these limitations, various methods for synthesizing light field images from normal images, {\em i.e.} without using a light field camera, have been proposed~\cite{ivan2019synthesizing,kalantari2016learning,mildenhall2019llff,srinivasan2019pushing,srinivasan2017learning,wing2018fast,wu2017light,zhou2018stereo}.
However, the goal is to synthesize light field still images, not videos.
To the best of our knowledge, no previous work has synthesized a light field video from a normal monocular video.
\begin{figure}[t]
	\begin{center}
		{\includegraphics[width=0.32\linewidth]{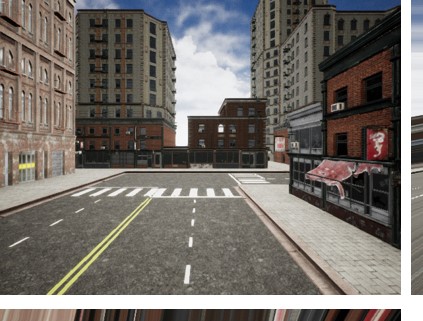}}~%
		{\includegraphics[width=0.33\linewidth]{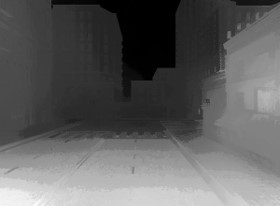}}~%
		{\includegraphics[width=0.33\linewidth]{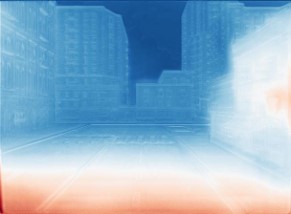}}~%
		
		\vspace{-2.0mm}
		\subfloat[]{\includegraphics[width=0.32\linewidth]{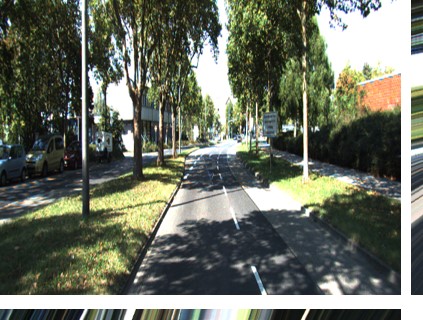}}~%
		\subfloat[]{\includegraphics[width=0.33\linewidth]{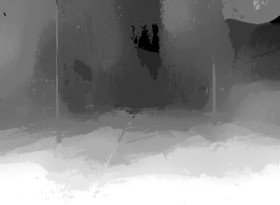}}~%
		\subfloat[]{\includegraphics[width=0.33\linewidth]{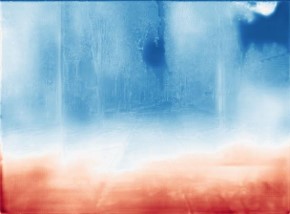}}~%
		\vspace{-4.0mm}
	\end{center}
	\caption{Synthesized light field frames and their applications. The network is trained only with the synthetic dataset. (a) Synthesized light field images (center sub-aperture images and epipolar images) on synthetic and actual datasets. (b) Estimated depth. (c) Estimated appearance flow.}
	\label{fig:Intro}
	\vspace{-4.0mm}
\end{figure}

Conventional learning-based light field synthesis methods are inspired by depth estimation techniques~\cite{flynn2016deepstereo,garg2016unsupervised,godard2017unsupervised,xie2016deep3d}.
Most methods require a sparse set (4$\sim$9) of input images to synthesize a single light field image ($8\times8 \sim 10\times10$).
\cite{srinivasan2017learning} synthesized a light field image from a single input image, but it is highly dependent on the quality of the estimated depth image.
\cite{ivan2019synthesizing} also synthesized a light field image from a single input image with improved generality of object and robustness.
Note that \cite{srinivasan2017learning} and \cite{ivan2019synthesizing} can only generate a light field video using a frame-by-frame approach.
Therefore, temporal consistency cannot be guaranteed.

A light field video capturing method was proposed by~\cite{wang2017light}.
It uses a hybrid camera system combining a general DSLR camera and a light field camera (Lytro) and synthesizes the light field video.
This method cannot be generalize to in-the-wild video capture and it is prone to error due to a viewpoint mismatch between two cameras.

Deep learning-based methods require a large-sized dataset to train the network.
However, acquiring a dataset that exactly fits a specific purpose is not trivial.
Especially for the light field, no video training dataset can be used for light field video processing.
To overcome this limitation, similar to the approach used in \cite{richter2016playing,zhang2017physically,richter2016playing,ros2016synthia,richter2017playing}, we use a graphics-generated photorealistic video simulated on a virtual environment.
In our approach, we use {\it UnrealCV}~\cite{qiu2016unrealcv} which is based on the Unreal4 game graphics engine to collect the synthetic light field video consisting of 9$\times$9 sub-aperture images (SAIs).
Using synthetic light field data avoids the limitations of previous light field images, {\em i.e.} low spatial/angular resolution, and small baseline.

In this paper, we propose a novel framework for 5D light field video synthesis.
We introduce a correlation layer to find correspondence between adjacent input frames and use it to estimate optical flow and appearance flow.
Figure~\ref{fig:Intro} shows the synthesized light field video frames and their applications for synthetic and actual scenes.
As shown in Figure~\ref{fig:Intro}, although we trained our framework with a synthetic dataset, it performs well in synthesizing a light field video from an actual scene.
The key contributions of this paper are summarized as follows:
\begin{itemize}
	\itemsep0.01em
	\item an end-to-end deep learning-based framework for 5D light field video synthesis.
	\item a new photorealistic synthetic dataset for light field video synthesis network training.
	\item capability of synthesizing light field video for actual images while the network is trained on the synthetic dataset.
\end{itemize}
\section{Related Works}
\paragraph{Light Field Synthesis}
The light field image, first proposed by Lippmann~\etal~\cite{lippmann1908epreuves}, was introduced in 2005 in the form of a plenoptic camera by Ng~\etal~\cite{ng2005light}, and its potential has since attracted attention.
At the same time, an increasing demand to synthesize light field images of a large amount of SAI from a small number of images is observed.
Wanner and Goldluecke~\cite{wanner2013variational} estimated the disparity maps using epipolar plane image (EPI) analysis of light field images and proposed super-resolution and view synthesis of light field images.
Zhang~\etal~\cite{zhang2015light} proposed a method for synthesizing light field images using a disparity assisted phase-based light field synthesis based on the difference of stereo images with a small baseline.

A learning-based light field image synthesis method was first proposed by Kalantari~\etal~\cite{kalantari2016learning}.
In this method, a network for estimating the disparity between each viewpoint and a network for correcting the color of the synthesized light field image are used to synthesize a 4D light field image using four input images located at the corners of the light field image.
Wu~\etal~\cite{wu2017light} synthesized the light field images from a minimum of nine and a maximum of 25 input images using an EPI super-resolution with a special blur kernel.
Wang~\etal~\cite{wang2018end} and Yeung~\etal~\cite{wing2018fast} used a 4D CNN for synthesizing light field images from a sparse set of input images.

A method for synthesizing light field image from a single image rather than multiple input images was first proposed by Srinivasan~\etal~\cite{srinivasan2017learning}.
In this method, a depth image corresponding to each SAI is initially estimated from an input image, and then a novel view is synthesized by warping the input image.
Then, a light field image is synthesized in the angular domain through a color estimation network.
However, the drawback is that the method depends heavily on the quality of the estimated depth and the color information of the image.
Ivan and Park~\cite{ivan2019synthesizing} synthesized the light field image using the appearance flow proposed by Zhou~\etal~\cite{zhou2016view} rather than the depth image.
However, as with~\cite{srinivasan2017learning}, the limitation is that it is unsuitable for synthesizing the light field video because it aims to synthesize the light field image of the static object.

A method for synthesizing a multiview image with a large baseline rather than a light field image has also been proposed.
Huang~\etal~\cite{huang2018deepmvs} proposed a method for synthesizing a horizontal light field image using multiple input images and camera pose information.
Zhou~\etal~\cite{zhou2018stereo} proposed a method for synthesizing a horizontal light field image from a stereo image with a small baseline using a new scene representation called a multiplane image (MPI).
However, these methods have limitations that require two or more input images or even camera parameters, which are difficult for a general user to provide.
In~\etal~\cite{srinivasan2019pushing}, \cite{zhou2018stereo} is extended, and a method for synthesizing light field images with a large baseline using MPI is proposed.
Mildenhall~\etal~\cite{mildenhall2019llff} proposed a method for synthesizing light field images with large baselines using MPI as well.
Although \cite{srinivasan2019pushing} and ~\cite{mildenhall2019llff} proposed a method for synthesizing a light field image with a large baseline rather than a conventional light field image, the limitation is that a large amount of computation and two or more input images are required.

The method for synthesizing light field videos was proposed by Wang~\etal~\cite{wang2017light} using a hybrid camera system consisting of a general DSLR camera and a light field camera.
It acquires videos simultaneously with a light field camera (3fps) and a DSLR camera (30fps), and then the light field video is synthesized.
However, this method requires not only a standard camera but also a light field camera.
In addition, errors occur due to the viewpoint mismatch between the two cameras.

In this paper, we deal with the synthesis of light field images, especially the synthesis of light field videos.
Different from conventional methods, no camera position information and no two or more input images are required to synthesize one light field image.
We also propose a deep learning-based framework that synthesizes a light field video from a monocular video rather than a single image for static objects.

\vspace*{-0.2cm}
\paragraph{Video Temporal Consistency}
Research has been conducted for a long time to solve the temporal inconsistency that occurs when an image processing method is applied to a video.
Bonnel~\etal~\cite{bonneel2015blind} proposed a method that provides temporal consistency to a video over general image processing methods, rather than specific image processing methods.
However, the limitation is that the corresponding method differs from the dense correspondence information used for each image processing method and depends heavily on the quality of that information.
Lai~\etal~\cite{lai2018learning} proposed a temporal consistency scheme for learning-based methods.
The method adds a long short term memory (LSTM) layer to various learning-based methods, such as colorization, enhancement, style transfer, and intrinsic decomposition of the image, to provide temporal consistency.
In this paper, we propose a method that provides temporal consistency for light field videos, not monocular videos.

\begin{figure}[t]
	\begin{center}
		\stackunder[5pt]{\includegraphics[width=\linewidth]{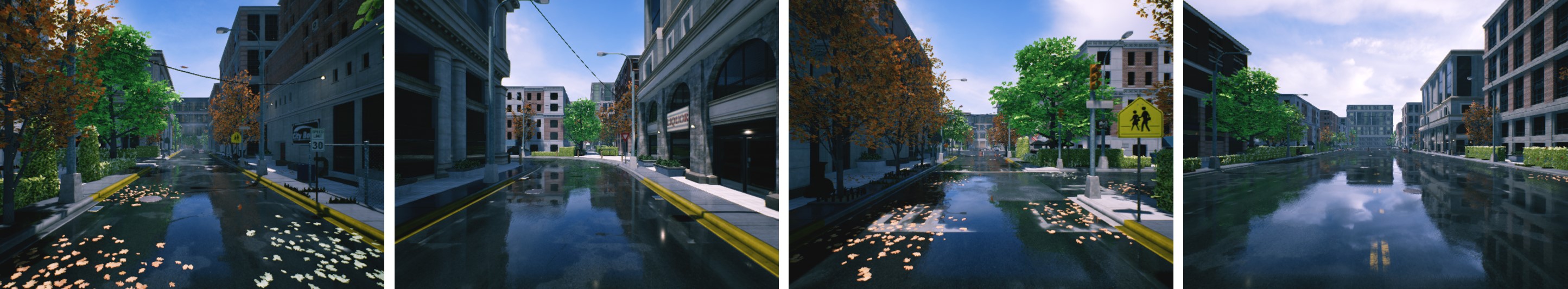}}~%
		\vspace{-1.0mm}
		\stackunder[5pt]{\includegraphics[width=\linewidth]{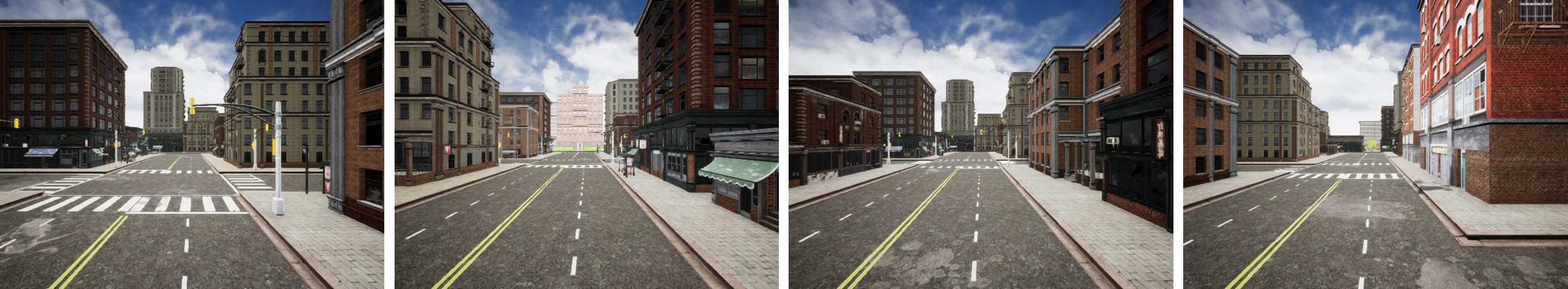}}~%
		\vspace{-1.0mm}
		\stackunder[5pt]{\includegraphics[width=\linewidth]{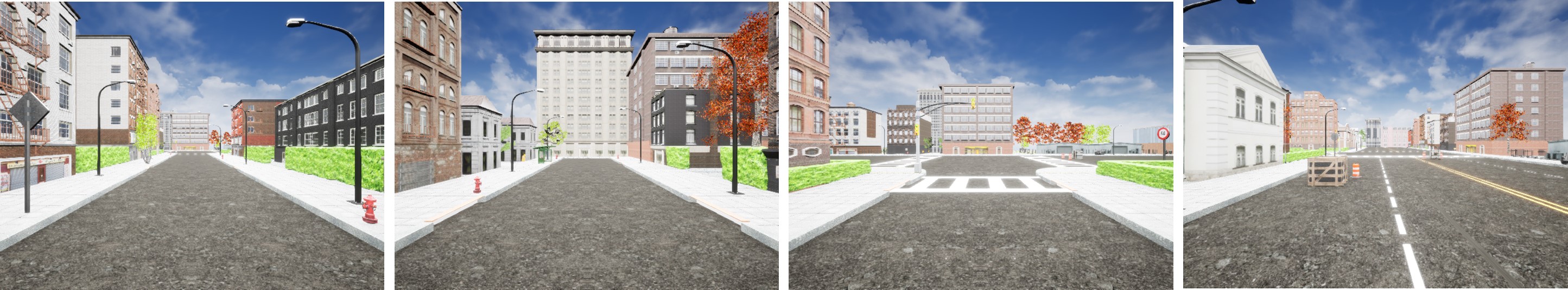}}~%
		\vspace{-4.0mm}
	\end{center}
	\caption{Examples of the synthetic dataset. The dataset consists of three different city environments, 24 scenes, and 3,333 frames.}
	\label{fig:dataset}
	\vspace{-3.0mm}
\end{figure}

\vspace*{-0.2cm}
\paragraph{Synthetic Dataset}
In the field of computer vision, many deep learning-based methods have been proposed.
However, deep learning networks for solving various computer vision problems require a large amount of dataset to train the network.
Many methods require a large amount of image data and a label of each image to train the network, thus consuming substantial time and effort.
To solve this problem, various methods that easily acquire data through synthetic environments have been proposed.
Richter~\etal~\cite{richter2016playing} proposed a method that easily acquires 3D scenes and labels by intervening in the communication process between the video game Grand Theft Auto \RNum{5} and the GPU.
Ros~\etal~\cite{ros2016synthia} proposed a method for acquiring 3D scene data by constructing a stereo camera on a virtual car after constructing a virtual city.
Zhang~\etal~\cite{zhang2017physically} used OpenGL for acquiring synthetic data.
However, these methods only acquire a single or stereo synthetic image, and the limitation is that a user's arbitrary scene is difficult to create.
Qiu~\etal~\cite{qiu2016unrealcv} proposed the {\it UnrealCV}, a module for acquiring user arbitrary scenes based on Unreal Engine 4, an open-source game engine that can easily construct user arbitrary scenes.
In this paper, we use UnrealCV to construct a virtual city and a light field camera and then acquire synthetic light field data as shown in Figure~\ref{fig:dataset}.

\begin{figure*}[t]
	\begin{center}
		{\includegraphics[width=\linewidth]{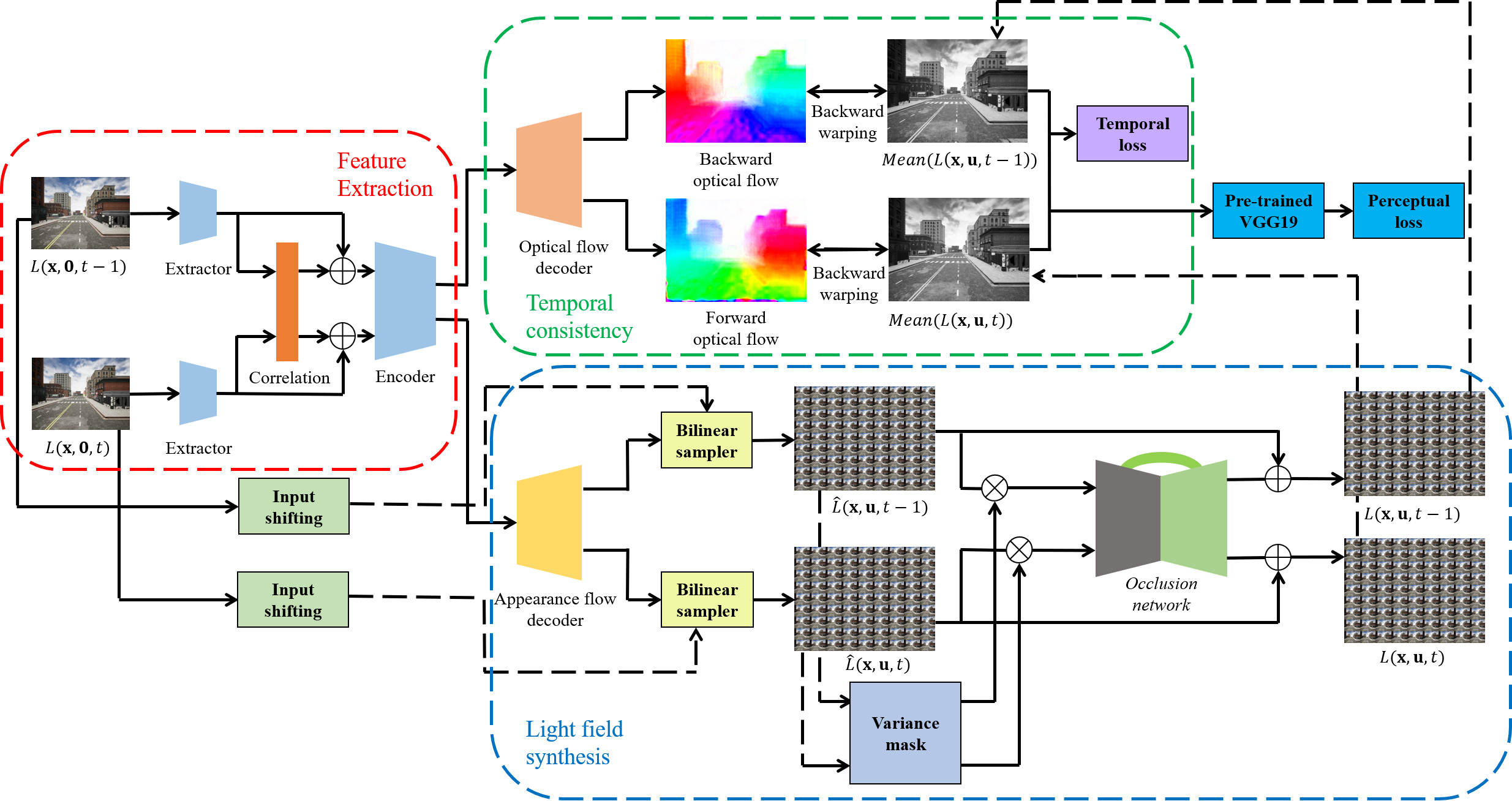}}~%
		\vspace{-4.0mm}
	\end{center}
	\caption{Proposed deep learning framework. The overall framework is divided into 3 parts, {\em i.e.} a feature extraction for extracting the features from each frame of the input video, 5D light field video synthesis, and temporal consistency refinement. After extracting the features by considering the correlation between adjacent frames, the light field video is synthesized by estimating both optical flow and appearance flow, and by refining using variance mask.}
	\label{fig:Network}
	\vspace{-4.0mm}
\end{figure*}
\section{Proposed Method}
In this paper, we propose a method for synthesizing 5D light field video  ${L}(\textbf{x},\textbf{u},t)$ from a monocular video ${L}(\textbf{x},\textbf{0},t)$.
We parameterize the light field video as ${L}(\textbf{x},\textbf{u},t)$ following~\cite{levoy1996light}.
$\textbf{x}$ and $\textbf {u}$ denote the coordinate vector in the spatial domain $(x, y)$ and the angular domain $(u, v)$ of the light field, respectively.
We represent the light field video synthesis as an approximation function $f(\cdot)$ as follows:
\begin{equation}\label{eq:1}
{L}(\textbf{x},\textbf{u},t) = f({L}(\textbf{x},\textbf{0},t)).
\end{equation}
Figure~\ref{fig:Network} shows the proposed deep learning framework that synthesizes 5D light field video from the input monocular video.
The overall framework is divided into three sections, namely, a feature extraction for extracting the features from each frame of the input video, 5D light field video synthesis, and temporal consistency refinement.
Since the ground truth appearance flow and optical flow are computationally expensive and difficult, the proposed framework estimates the appearance flow and optical flow by training through an unsupervised approach.

\subsection{Feature Extraction}\label{sec:3-1}
${L}(\textbf{x},\textbf{0},t-1)$, and ${L}(\textbf{x},\textbf{0},t)$ are initially converted into the luminance images for memory efficiency to extract the feature from the input monocular video frame.
The initial feature~$\xi_t^{(l)}$ is extracted using an initial feature extraction encoder~$\phi_{init}$, which consists of four convolution layers.
The process of extracting the initial feature can be described as follows:
\begin{equation}\label{eq:2}
\xi_t^{(l)} = \phi_{init}^{(l)}({L}(\textbf{x},\textbf{0},t))
\end{equation}
where $\phi_{init}^{(l)}(\cdot)$ represents feature activation in the $l_{\rm th}$ layer of the initial feature extraction encoder~$\phi_{init}$.
The extracted initial features are passed through the correlation layer in consideration of the correlation between two adjacent video frames, namely,~${L}(\textbf{x},\textbf{0},t-1)$ and ${L}(\textbf{x},\textbf{0},t)$.
The encoder~$\phi_{fin}$ extracts the final feature~$\zeta_t$ and $\zeta_{t-1}$ by combining the obtained correlation information and the initial features.
The process of extracting the final feature can be described as follows:
\begin{align}
\zeta_t = \phi_{fin}(corr(\xi_t^{(4)}, \xi_{t-1}^{(4)}), conv(\xi_t^{(4)})) \label{eq:3} \\ 
\zeta_{t-1} = \phi_{fin}(corr(\xi_{t-1}^{(4)}, \xi_{t}^{(4)}), conv(\xi_{t-1}^{(4)})) \label{eq:4}
\end{align}
where $corr$ and $conv$ represent the correlation and convolution layers, respectively.

\subsection{Light Field Synthesis}
\paragraph{Appearance Flow Estimation}
The appearance flow decoder~$\varphi_f$, which takes the encoded feature map $\zeta_t$ obtained from Eq.~(\ref{eq:3}), estimates the appearance flow~${L}_f(\textbf{x},\textbf{u},t)$ corresponding to each SAI at time $t$.
Using estimated appearance flow, we synthesize the initial light field video frame $\hat{L}(\textbf{x},\textbf{u},t)$ by warping shifted input images~${L}_s(\textbf{x},\textbf{u},t)$ and estimated appearance flow for each angular coordinate.
The initial light field synthesis can be written as follows:

\begin{align}
\hat{L}(\textbf{x},\textbf{u},t)&=B({L}_s(\textbf{x},\textbf{u},t), {L}_f(\textbf{x},\textbf{u},t))\label{eq:5} \\
{L}_f(\textbf{x},\textbf{u},t)&=\varphi_f(\zeta_t)\label{eq:6} \\
{L}_s(\textbf{x},\textbf{u},t)&= S({L}(\textbf{x},\textbf{0},t), \nabla(\textbf{u}))\nonumber \\
&= {L}(x-(\eta\times\Delta{u}), y-(\eta\times\Delta{v}))\label{eq:7}
\end{align}
where $B$ is the warping function for synthesizing the light field using shifted input image ${L}_s(\textbf{x},\textbf{u},t)$ and its corresponding appearance flow ${L}_f(\textbf{x},\textbf{u},t)$.
The warping function uses the bilinear sampler module~\cite{jaderberg2015spatial} for generating initial light field video frame $\hat{L}(\textbf{x},\textbf{u},t)$.
$S$ is the input shifting technique, which shifts the central view to the position $\nabla(\textbf{u})$ of the novel view.
The input shifting technique eases the difficulty of training the network by working as the bias initialization.
$\eta$ is a shift constant based on the disparity between each SAI.
We use the light field mean and variance losses~$\ell_{local}$, $\ell_{global}$ proposed by \cite{ivan2019synthesizing} for training the appearance flow decoder.

\begin{figure}[t]
	\begin{center}
		\subfloat[]{\includegraphics[width=0.33\linewidth]{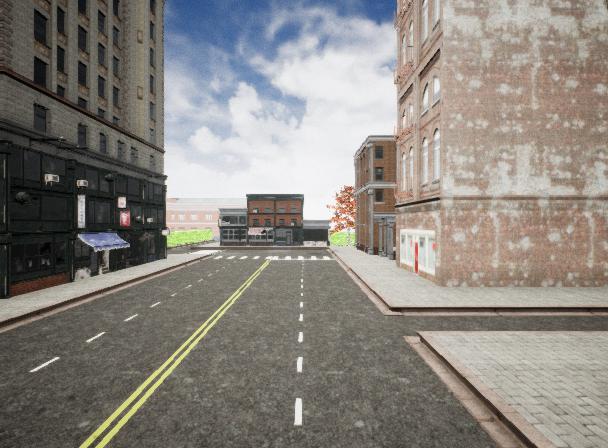}}~%
		\subfloat[]{\includegraphics[width=0.33\linewidth]{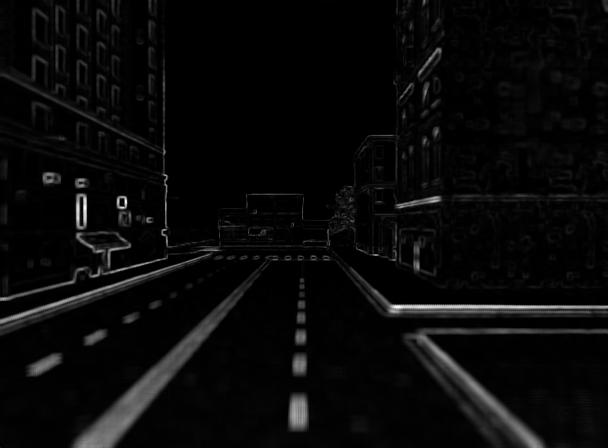}}~%
		\subfloat[]{\includegraphics[width=0.33\linewidth]{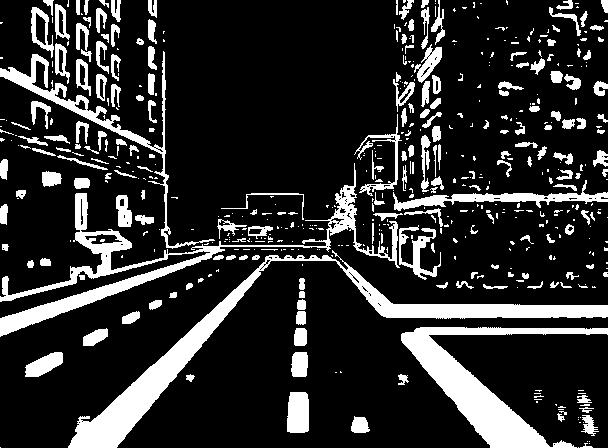}}~%
	\end{center}
	\vspace{-4.0mm}
	\caption{Light field variance mask. (a) Centerview. (b) Variance image of light field. (c) Variance mask.}
	\label{fig:variance}
	\vspace{-3.0mm}
\end{figure}

\paragraph{Occlusion Network}
The synthesized light field video frame obtained by Eq.~(\ref{eq:5}) has a limitation, {\em i.e.}, the synthesis result of the image boundary and occluded region is poor due to the nature of appearance flow.
As shown in Figure~\ref{fig:variance}, a 9$\times$9 variance mask~$Mask(\hat{L})$ is formed using $\hat{L}(\textbf{x},\textbf{u},t)$'s variance image~$Var(\hat{L})$ to overcome this limitation.
The variance image of the light field represents a difference between each SAI, which is the occluded region of the scene.
To improve the quality of the occluded, edged, and boundary regions of the synthesized light field video frame, we propose {\em Occlusion Network}, which improves the quality of the synthesized light field video frame by inputting the synthesized light field video frame and the variance mask.
Figure~\ref{fig:variance} visualizes the variance image and mask.
Different from the 2D image, the light field image has an angular dimension in addition to the spatial dimension.
To handle both dimensions simultaneously, {\em Occlusion Network} is constructed using the 3D convolution layers instead of the	2D convolution layers.
{\em Occlusion Network} has a 3D encoder-decoder structure with $\hat{L}(\textbf{x},\textbf{u},t)$ and $Var(\hat{L})$ as inputs.
To preserve the information of the angular dimension, {\em Occlusion Network} maintains the size of the angular dimension while reducing the spatial resolution when it passes through the encoder layers.
By contrast, the decoder layer increases the spatial resolution while maintaining the size of the angular dimension.
The final light field video frame~${L}(\textbf{x},\textbf{u},t)$ is obtained by adding~$\hat{L}(\textbf{x},\textbf{u},t)$ and the residual image~$R(\textbf{x},\textbf{u},t)$ obtained at the last stage of the decoder layer, which can be described as follows:
\begin{flalign}
{L}(\textbf{x},\textbf{u},t) = O(\hat{L}(\textbf{x},\textbf{u},t), Var(\hat{L})) + \hat{L}(\textbf{x},\textbf{u},t)\label{eq:8}
\end{flalign}
where $O(\cdot)$ represents the {\em Occlusion Network}.
We train the {\em Occlusion Network} by minimizing the simple L1 error as follows:
\begin{flalign}\label{eq:9}
\ell_{occ} = ||{L}(\textbf{x},\textbf{u},t) - {L_{GT}}(\textbf{x},\textbf{u},t)||_1.
\end{flalign}
The 3D convolution is performed with the filter size 3$\times$3$\times$3, and leaky Relu~\cite{xu2015empirical} activation function.
The last layer uses tanh activation function to force the pixel values of the residual image to [-1, 1].
\paragraph{Perceptual Loss}
To improve the synthesized light field video frame's quality, we use perceptual loss using the pre-trained VGG-19 network~\cite{simonyan2014very}.
Perceptual loss has been used as a loss function in various deep learning-based methods; it has a similar tendency to human perception~\cite{zhang2018unreasonable}.
Computing perceptual loss for all SAIs of light field video frames is computationally heavy.
Therefore, we calculate and optimize the perceptual loss for the mean image of the synthesized light field video frame.
Perceptual loss can be described as follows:
\begin{flalign}\label{eq:10}
\ell_{percep} = ||&\phi_{VGG}^{(3)}(Mean(L(\textbf{x},\textbf{u},t))) - \\\nonumber
&\phi_{VGG}^{(3)}(Mean(L_{GT}(\textbf{x},\textbf{u},t)))||_1
\end{flalign}
where $\phi_{VGG}^{(n)}(\cdot)$ represents the $n_{\rm th}$ feature activation of the VGG-19 network.
We use the third layer's feature activation which contains low-level features to calculate the perceptual loss.

\subsection{Temporal Consistency}\label{sub:3-2-4}
In general, an optical flow between each frame is estimated and warped to a target frame to minimize the difference between the warped frame and target frame and provide temporal consistency to a video.
However, estimating the optical flow for every 9$\times$9 SAIs in the light field video frame is difficult.
Therefore, we do not estimate the optical flow for all 9$\times$9 SAIs, but we estimate the optical flow for the mean image of the light field video frame~$Mean(\hat{L})$ and use it to maintain temporal consistency.
To estimate the optical flow, we use the features extracted from $\phi_{init}$ and $\phi_{fin}$ and the loss function in~\cite{meister2018unflow}.
To obtain the accurate optical flow for the light field image, we estimate the optical flow between the input video frames ${L}(\textbf{x},\textbf{0},t)$ and ${L}(\textbf{x},\textbf{0},t-1)$.
After the network that synthesizes the initial light field image is trained to some level, the two synthesized initial light field video frames~$\hat{L}(\textbf{x},\textbf{u},t)$ and~$\hat{L}(\textbf{x},\textbf{u},t-1)$ are used instead of the two input video frames.
Using the mean image of the light fields $\hat{L}(\textbf{x},\textbf{u},t)$ and $\hat{L}(\textbf{x},\textbf{u},t-1)$, we estimate and warp the optical flow to ${(t-1) \rightarrow t}$ and ${t \rightarrow (t-1)}$.
The temporal consistency is obtained by minimizing the difference between the mean image of the warped image and the target light field image as follows:
\begin{flalign}\label{eq:3.14}
\ell_{temp} = ||&M(\omega(Mean({L}(\textbf{x},\textbf{u},t)), {O}_{t \rightarrow t-1})) - \\\nonumber
&M(Mean({L}(\textbf{x},\textbf{u},t-1)))||_1 + \\\nonumber
||&M(\omega(Mean({L}(\textbf{x},\textbf{u},t-1)), {O}_{t-1 \rightarrow t})) - \\\nonumber
&M(Mean({L}(\textbf{x},\textbf{u},t)))||_1
\end{flalign}
where $M$ is the valid mask generated using forward and backward optical flow as proposed by~\cite{zou2018df}.
The valid mask uses the assumption that the vectors must be in the same position when the vectors of the estimated optical flow is forwarded and then moved backward.
If the pixel's location satisfies the hypothesis, the pixel value is set to 1, and the other value is set to 0 to form a valid mask.

\section{Experimental Results}
We evaluate the proposed method qualitatively and quantitatively.
The performance is compared with the state-of-the-art methods for single image light field synthesis~\cite{srinivasan2017learning} and \cite{ivan2019synthesizing} because no prior work was conducted on light field video synthesis.
Moreover, no ground truth light field video dataset is avaliable.
Thus, we use the synthetic light field video dataset for the quantitative evaluation.
Test data consist of 162 frames that are unused in the training.
The qualitative evaluation is performed on the KITTI~\cite{Geiger2012CVPR} dataset, which is an actual scene dataset.
The spatial resolution of the training light field video data used in this paper is 480$\times$640, and the resolution of the original KITTI data is 375$\times$1242.
Therefore, the light field is synthesized by changing the input resolution to 480$\times$640 using the bicubic interpolation.
The proposed framework is implemented with TensorFlow~\cite{abadi2016tensorflow}.
While training, we randomly crop the input light field spatial resolution into 224$\times$224.

We train our network end-to-end on NVIDIA Titan RTX D6 24GB GPU, 16GB RAM, and Intel Core i9-9900X CPU @3.50GHz CPU using the Adam optimizer with default parameter $\beta_{1}=0.9$, $\beta_{2}=0.999$ and learning rate $\alpha=0.0002$.
The optical flow decoder and appearance flow decoder run for the first 50K iterations, and then every sub-networks run together.
More technical details are described in the supplementary material.

\paragraph{Temporal Stability}
To evaluate the temporal stability of the synthesized light field video, we calculate the temporal stability with optical flow-based warping error based on~\cite{lai2018learning} as follows:
\begin{flalign}\label{eq:4.1}
&E_{temp}(L(\textbf{x},\textbf{u},t), L(\textbf{x},\textbf{u},t-1))&\\\nonumber
&= \frac{1}{UV-1}\displaystyle\sum_{\textbf{u}=\textbf{-4}}^{\textbf{4}}\left[\frac{||\tilde{L}(\textbf{x},\textbf{u},t) - L(\textbf{x},\textbf{u},t-1) ||_1}{\displaystyle\sum_{n=1}^{N} M^{(n)}}\right], \textbf{u} \neq \textbf{0}
\end{flalign}
where $U$, and $V$ denote the size of each angular dimension and $M^{(n)}$ denotes the $nth$ pixel that has a value of 1 in the valid mask among $N$ pixels.
$\tilde{L}(\textbf{x},\textbf{u},t)$ denotes the frame warped $L(\textbf{x},\textbf{u},t)$ at time $t-1$.

\subsection{Qualitative Evaluation}\label{sec:4-2}
\begin{figure*}[t]
	\begin{center}
		\captionsetup[subfigure]{labelformat=empty}
		\subfloat[]{\raisebox{5.5mm}{\rotatebox[origin=t]{90}{Input}}}~%
		\subfloat[]{\includegraphics[width=0.115\linewidth]{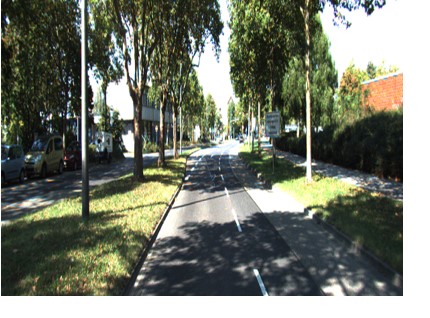}}~%
		\subfloat[]{\includegraphics[width=0.115\linewidth]{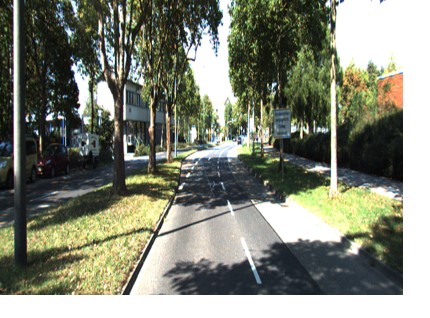}}~%
		\subfloat[]{\includegraphics[width=0.115\linewidth]{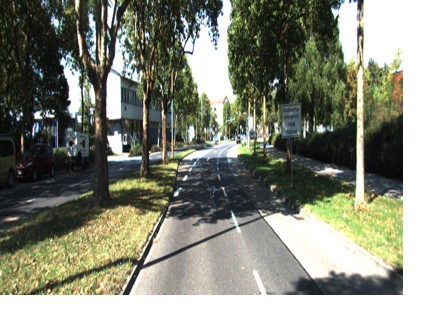}}~%
		\subfloat[]{\includegraphics[width=0.115\linewidth]{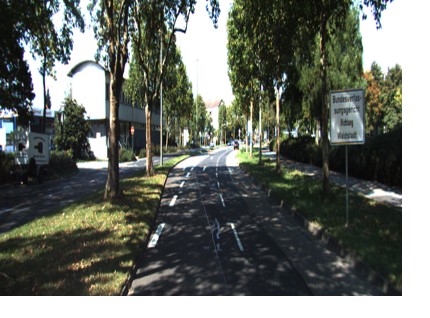}}~%
		\subfloat[]{\includegraphics[width=0.115\linewidth]{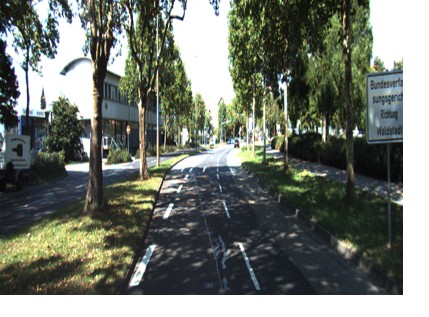}}~%
		\subfloat[]{\includegraphics[width=0.115\linewidth]{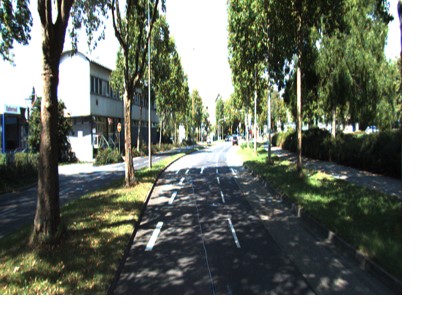}}~%
		\subfloat[]{\includegraphics[width=0.115\linewidth]{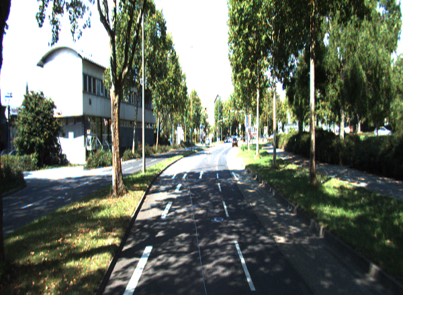}}~%
		\subfloat[]{\includegraphics[width=0.115\linewidth]{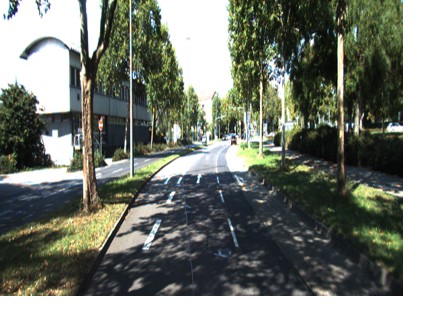}}~%
		
		\vspace{-7.0mm}
		\subfloat[]{\raisebox{5.5mm}{\rotatebox[origin=t]{90}{\cite{srinivasan2017learning}}}}~%
		\subfloat[]{\includegraphics[width=0.115\linewidth]{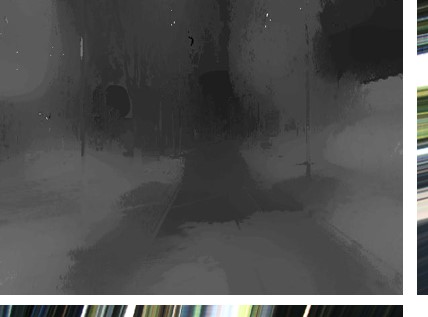}}~%
		\subfloat[]{\includegraphics[width=0.115\linewidth]{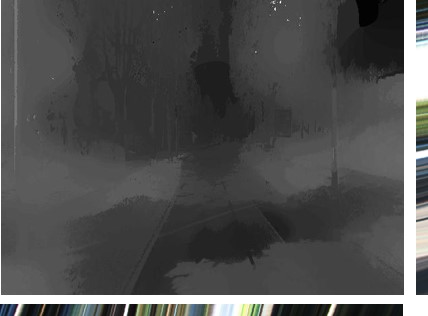}}~%
		\subfloat[]{\includegraphics[width=0.115\linewidth]{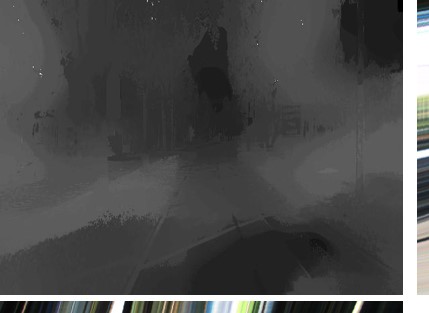}}~%
		\subfloat[]{\includegraphics[width=0.115\linewidth]{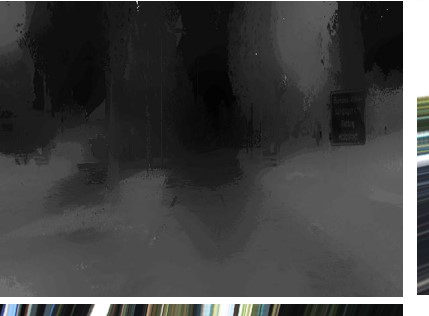}}~%
		\subfloat[]{\includegraphics[width=0.115\linewidth]{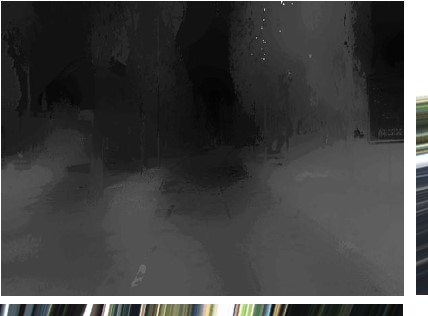}}~%
		\subfloat[]{\includegraphics[width=0.115\linewidth]{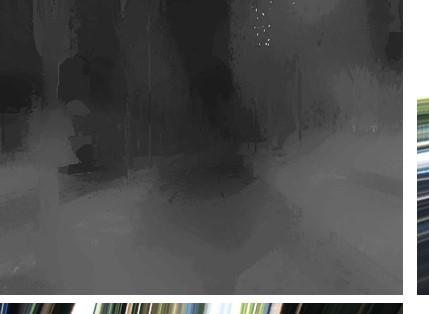}}~%
		\subfloat[]{\includegraphics[width=0.115\linewidth]{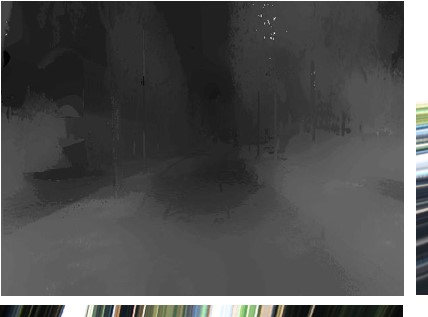}}~%
		\subfloat[]{\includegraphics[width=0.115\linewidth]{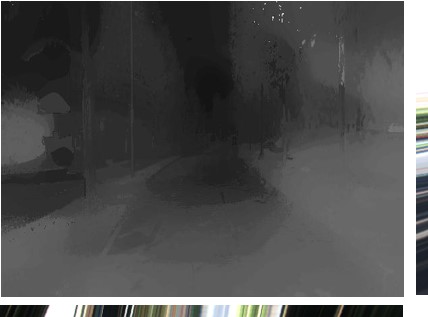}}~%
		
		\vspace{-7.0mm}
		\subfloat[]{\raisebox{5.5mm}{\rotatebox[origin=t]{90}{\cite{ivan2019synthesizing}}}}~%
		\subfloat[]{\includegraphics[width=0.115\linewidth]{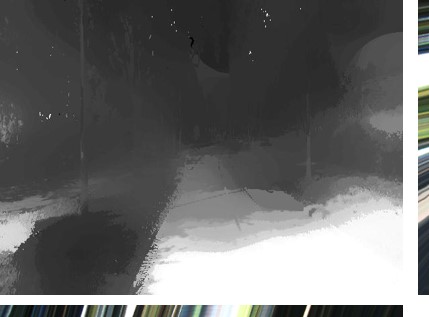}}~%
		\subfloat[]{\includegraphics[width=0.115\linewidth]{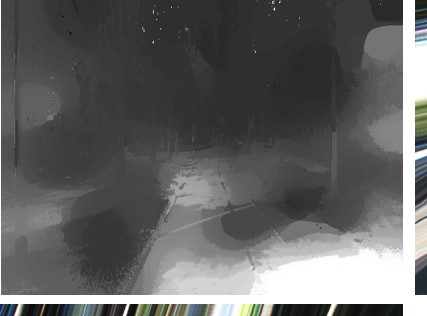}}~%
		\subfloat[]{\includegraphics[width=0.115\linewidth]{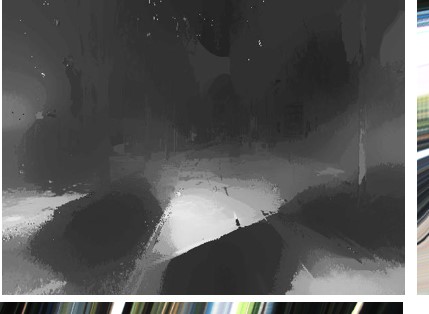}}~%
		\subfloat[]{\includegraphics[width=0.115\linewidth]{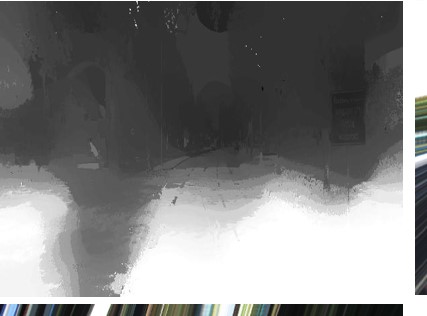}}~%
		\subfloat[]{\includegraphics[width=0.115\linewidth]{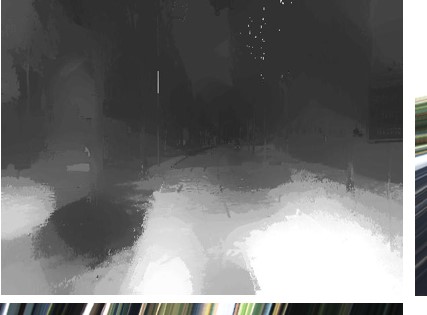}}~%
		\subfloat[]{\includegraphics[width=0.115\linewidth]{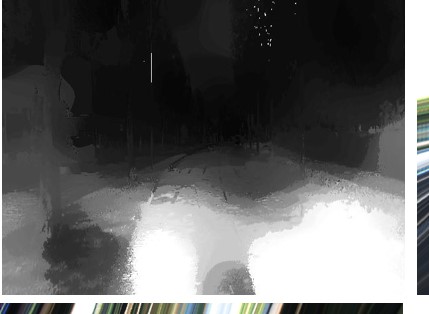}}~%
		\subfloat[]{\includegraphics[width=0.115\linewidth]{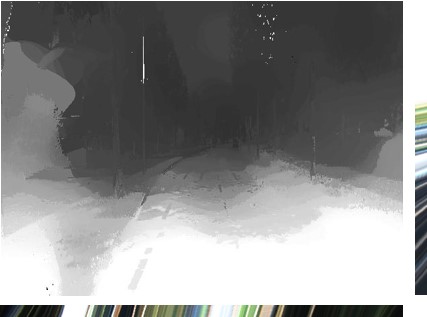}}~%
		\subfloat[]{\includegraphics[width=0.115\linewidth]{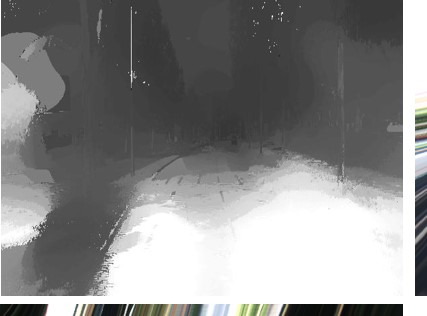}}~%
		
		\vspace{-7.0mm}
		\subfloat[]{\raisebox{5.5mm}{\rotatebox[origin=t]{90}{Proposed}}}~%
		\subfloat[]{\includegraphics[width=0.115\linewidth]{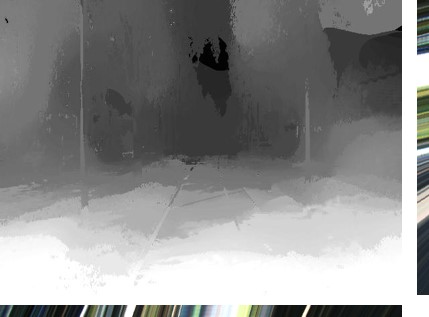}}~%
		\subfloat[]{\includegraphics[width=0.115\linewidth]{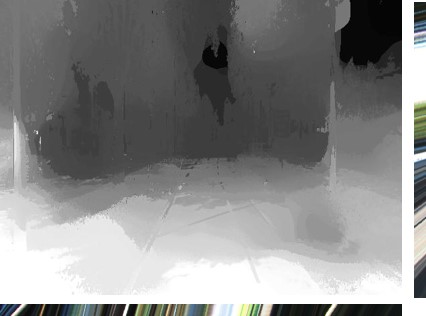}}~%
		\subfloat[]{\includegraphics[width=0.115\linewidth]{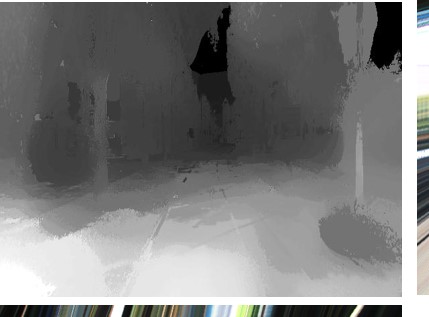}}~%
		\subfloat[]{\includegraphics[width=0.115\linewidth]{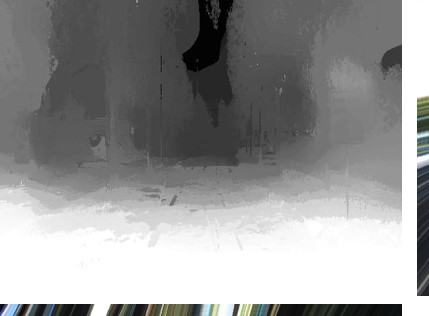}}~%
		\subfloat[]{\includegraphics[width=0.115\linewidth]{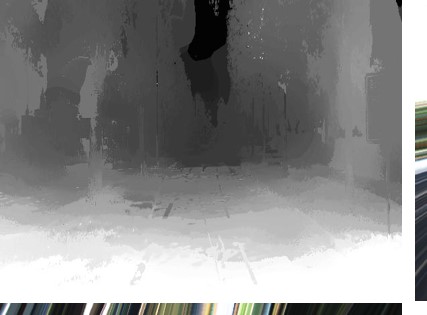}}~%
		\subfloat[]{\includegraphics[width=0.115\linewidth]{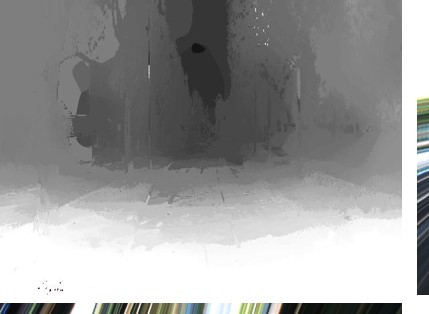}}~%
		\subfloat[]{\includegraphics[width=0.115\linewidth]{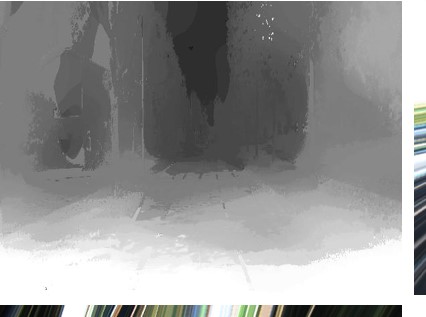}}~%
		\subfloat[]{\includegraphics[width=0.115\linewidth]{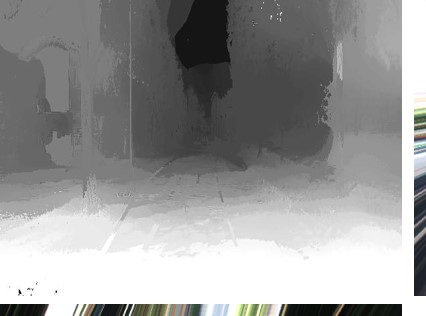}}~%
		
		\vspace{-4.0mm}
		\subfloat[]{\raisebox{5.5mm}{\rotatebox[origin=t]{90}{Input}}}~%
		\subfloat[]{\includegraphics[width=0.115\linewidth]{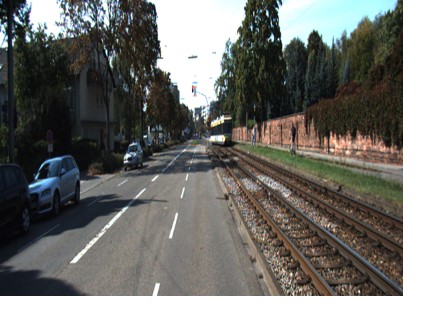}}~%
		\subfloat[]{\includegraphics[width=0.115\linewidth]{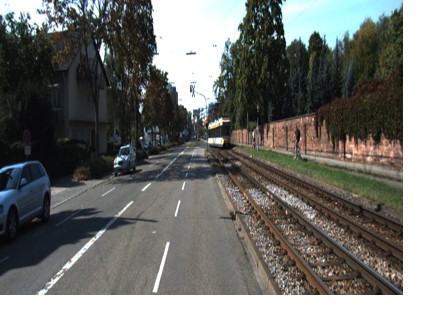}}~%
		\subfloat[]{\includegraphics[width=0.115\linewidth]{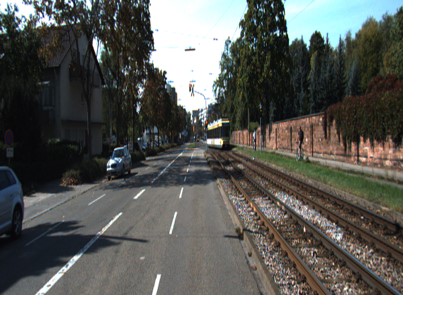}}~%
		\subfloat[]{\includegraphics[width=0.115\linewidth]{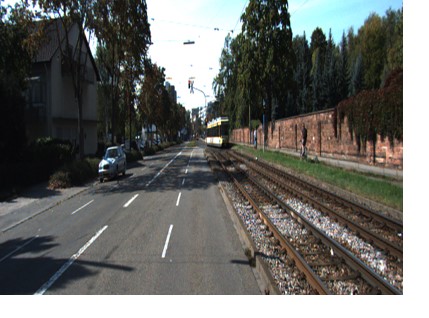}}~%
		\subfloat[]{\includegraphics[width=0.115\linewidth]{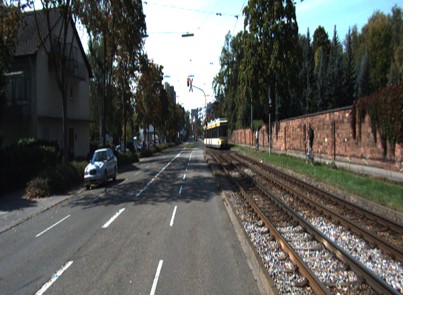}}~%
		\subfloat[]{\includegraphics[width=0.115\linewidth]{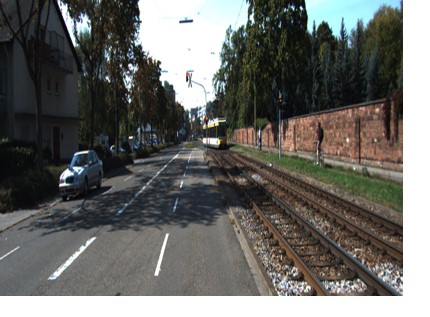}}~%
		\subfloat[]{\includegraphics[width=0.115\linewidth]{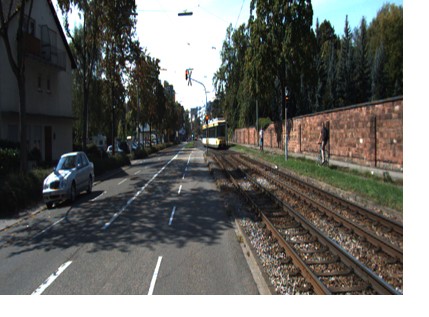}}~%
		\subfloat[]{\includegraphics[width=0.115\linewidth]{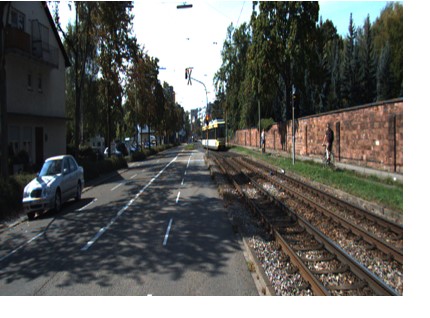}}~%
		
		\vspace{-7.0mm}
		\subfloat[]{\raisebox{5.5mm}{\rotatebox[origin=t]{90}{\cite{srinivasan2017learning}}}}~%
		\subfloat[]{\includegraphics[width=0.115\linewidth]{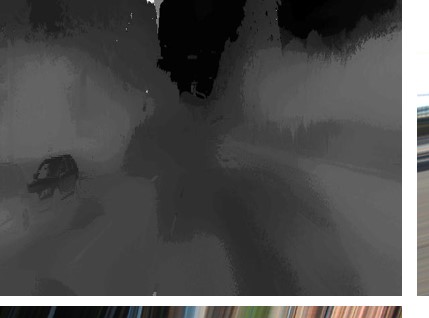}}~%
		\subfloat[]{\includegraphics[width=0.115\linewidth]{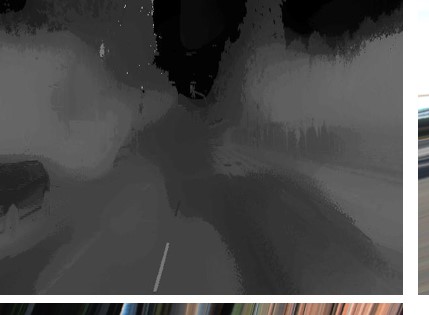}}~%
		\subfloat[]{\includegraphics[width=0.115\linewidth]{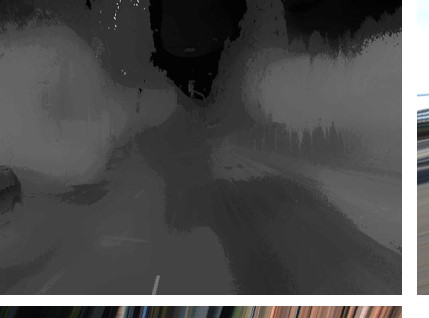}}~%
		\subfloat[]{\includegraphics[width=0.115\linewidth]{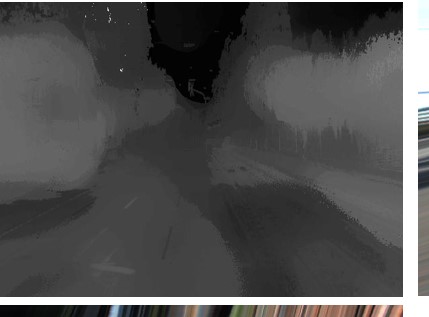}}~%
		\subfloat[]{\includegraphics[width=0.115\linewidth]{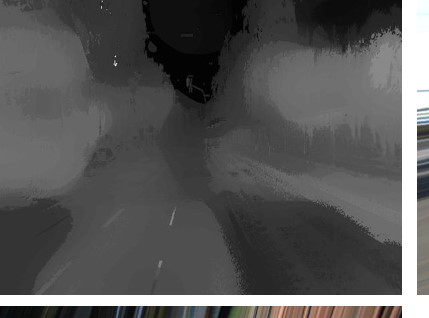}}~%
		\subfloat[]{\includegraphics[width=0.115\linewidth]{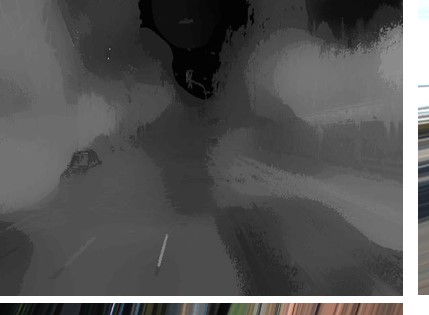}}~%
		\subfloat[]{\includegraphics[width=0.115\linewidth]{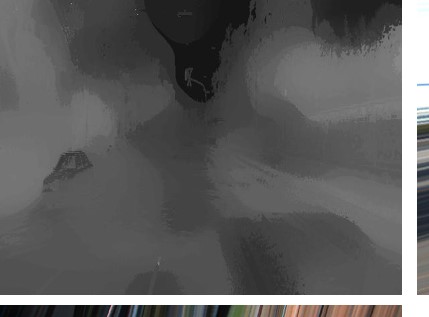}}~%
		\subfloat[]{\includegraphics[width=0.115\linewidth]{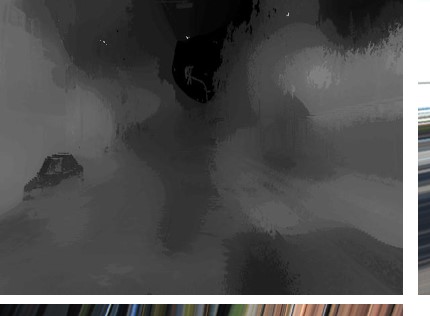}}~%
		
		\vspace{-7.0mm}
		\subfloat[]{\raisebox{5.5mm}{\rotatebox[origin=t]{90}{\cite{ivan2019synthesizing}}}}~%
		\subfloat[]{\includegraphics[width=0.115\linewidth]{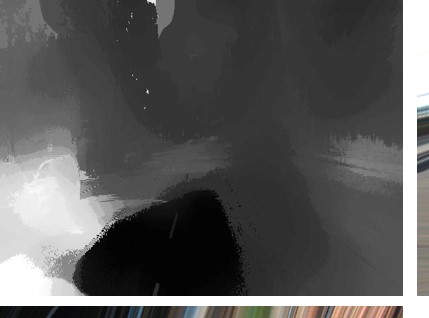}}~%
		\subfloat[]{\includegraphics[width=0.115\linewidth]{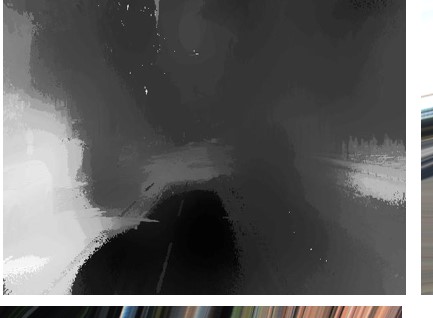}}~%
		\subfloat[]{\includegraphics[width=0.115\linewidth]{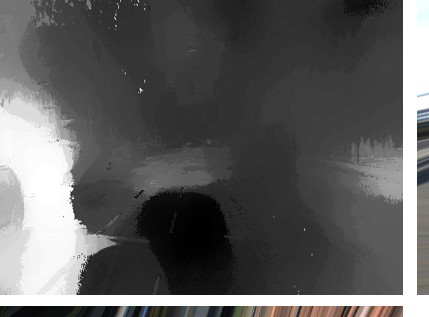}}~%
		\subfloat[]{\includegraphics[width=0.115\linewidth]{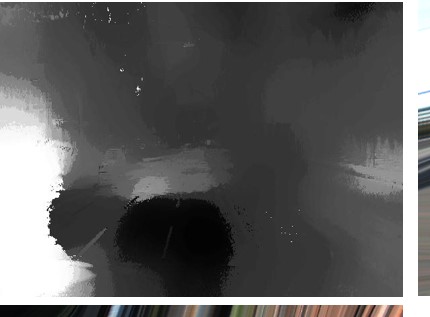}}~%
		\subfloat[]{\includegraphics[width=0.115\linewidth]{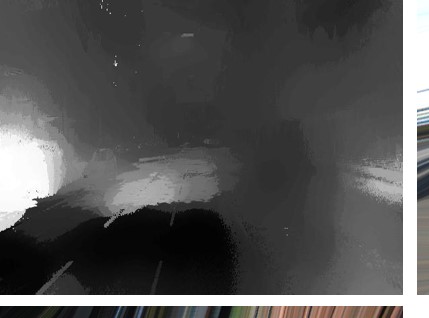}}~%
		\subfloat[]{\includegraphics[width=0.115\linewidth]{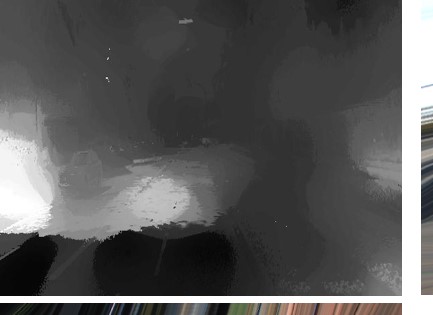}}~%
		\subfloat[]{\includegraphics[width=0.115\linewidth]{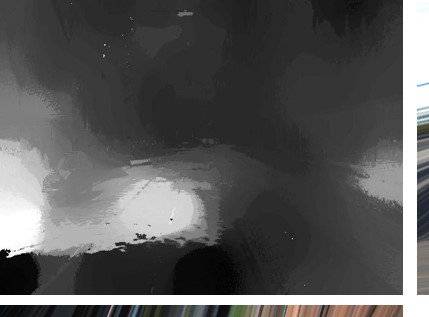}}~%
		\subfloat[]{\includegraphics[width=0.115\linewidth]{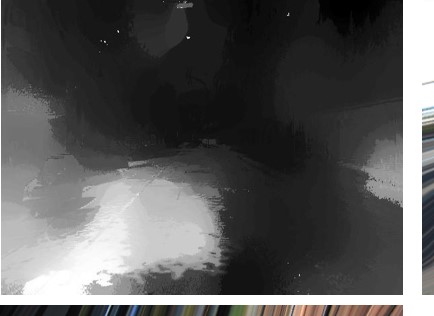}}~%
		
		\vspace{-7.0mm}
		\subfloat[]{\raisebox{5.0mm}{\rotatebox[origin=t]{90}{Proposed}}}~%
		\subfloat[]{\includegraphics[width=0.115\linewidth]{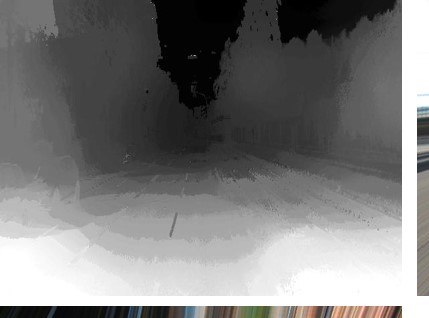}}~%
		\subfloat[]{\includegraphics[width=0.115\linewidth]{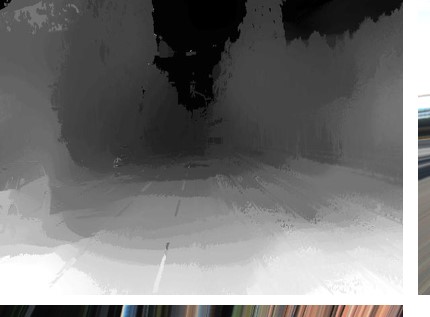}}~%
		\subfloat[]{\includegraphics[width=0.115\linewidth]{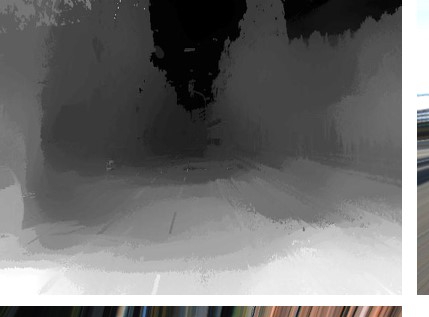}}~%
		\subfloat[]{\includegraphics[width=0.115\linewidth]{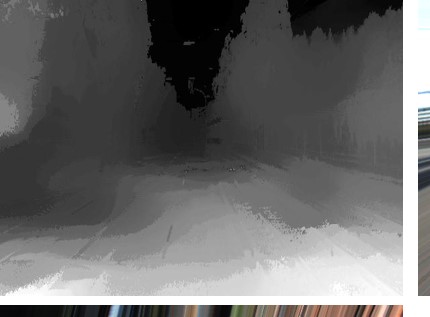}}~%
		\subfloat[]{\includegraphics[width=0.115\linewidth]{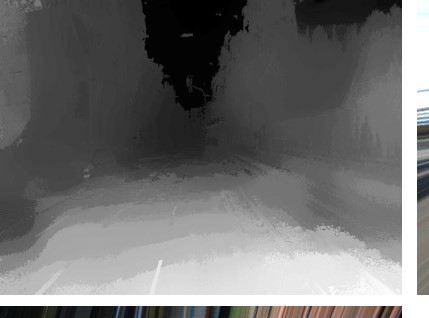}}~%
		\subfloat[]{\includegraphics[width=0.115\linewidth]{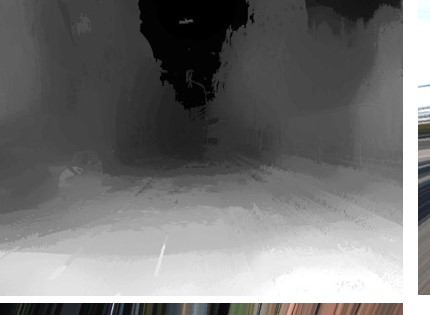}}~%
		\subfloat[]{\includegraphics[width=0.115\linewidth]{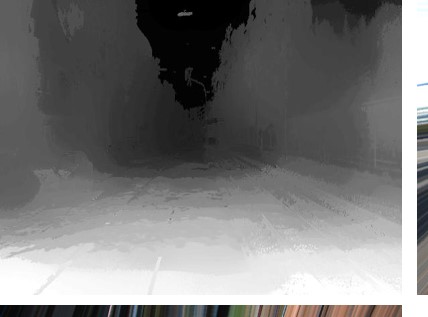}}~%
		\subfloat[]{\includegraphics[width=0.115\linewidth]{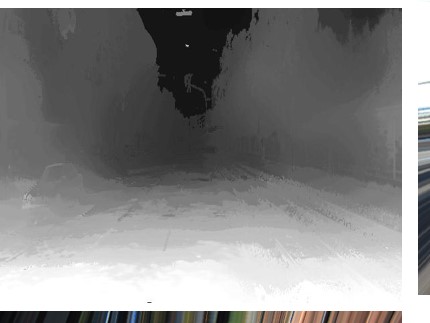}}~%
		
		\vspace{-4.0mm}
		\subfloat[]{\raisebox{5.5mm}{\rotatebox[origin=t]{90}{Input}}}~%
		\subfloat[]{\includegraphics[width=0.115\linewidth]{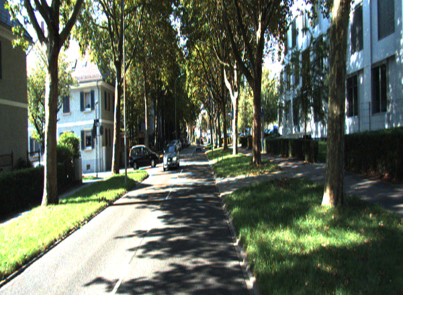}}~%
		\subfloat[]{\includegraphics[width=0.115\linewidth]{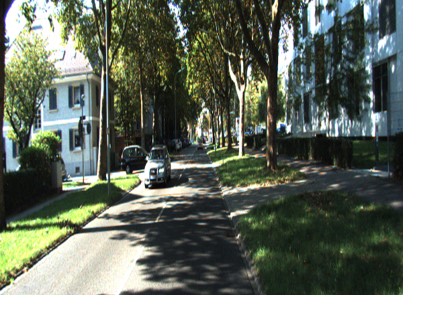}}~%
		\subfloat[]{\includegraphics[width=0.115\linewidth]{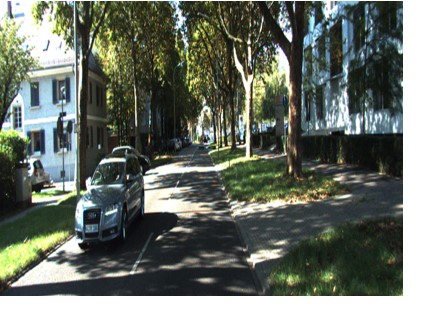}}~%
		\subfloat[]{\includegraphics[width=0.115\linewidth]{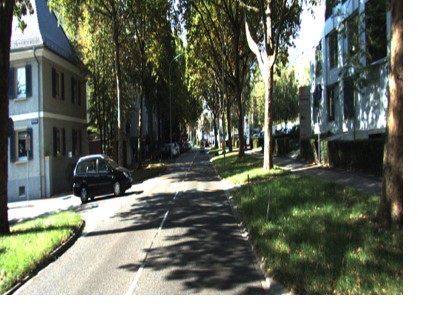}}~%
		\subfloat[]{\includegraphics[width=0.115\linewidth]{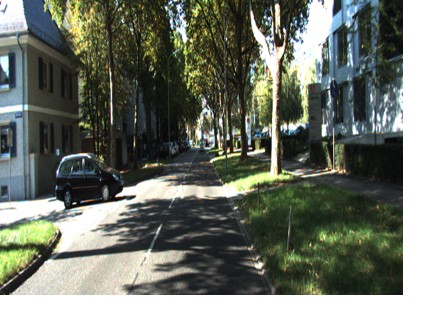}}~%
		\subfloat[]{\includegraphics[width=0.115\linewidth]{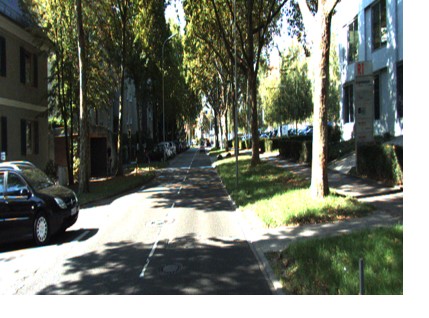}}~%
		\subfloat[]{\includegraphics[width=0.115\linewidth]{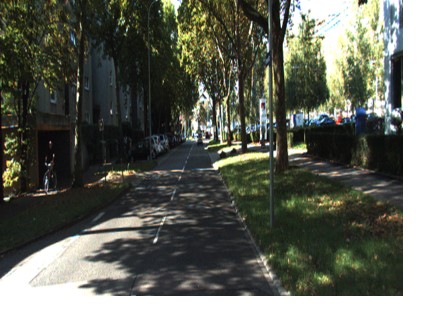}}~%
		\subfloat[]{\includegraphics[width=0.115\linewidth]{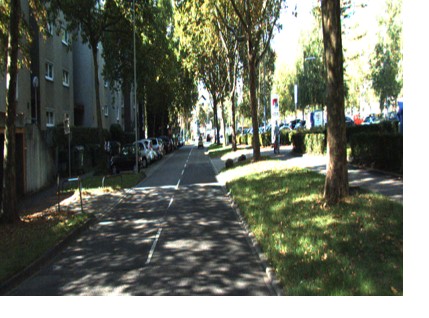}}~%
		
		\vspace{-7.0mm}
		\subfloat[]{\raisebox{5.5mm}{\rotatebox[origin=t]{90}{\cite{srinivasan2017learning}}}}~%
		\subfloat[]{\includegraphics[width=0.115\linewidth]{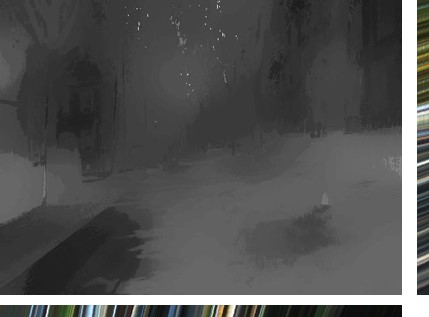}}~%
		\subfloat[]{\includegraphics[width=0.115\linewidth]{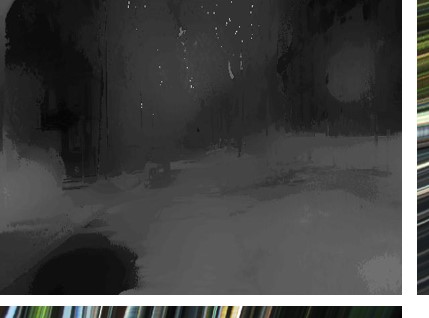}}~%
		\subfloat[]{\includegraphics[width=0.115\linewidth]{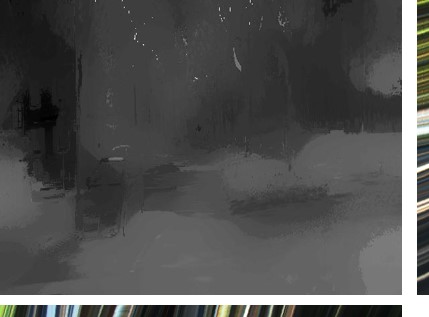}}~%
		\subfloat[]{\includegraphics[width=0.115\linewidth]{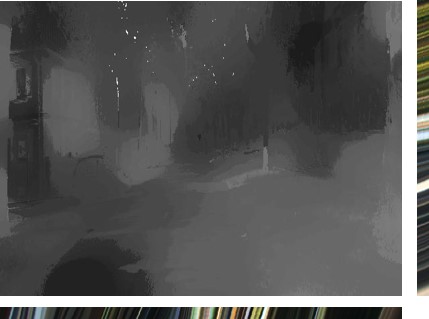}}~%
		\subfloat[]{\includegraphics[width=0.115\linewidth]{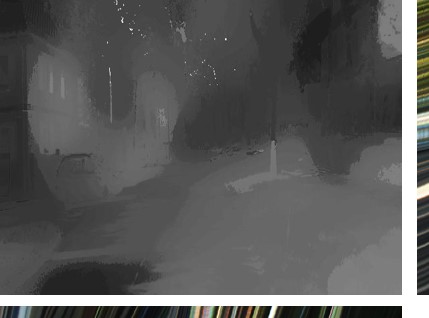}}~%
		\subfloat[]{\includegraphics[width=0.115\linewidth]{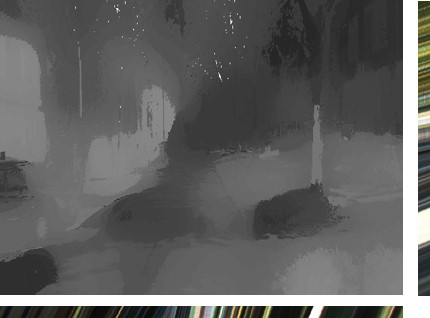}}~%
		\subfloat[]{\includegraphics[width=0.115\linewidth]{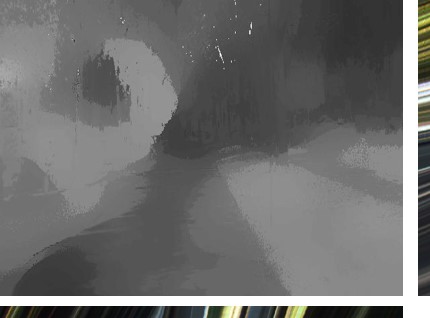}}~%
		\subfloat[]{\includegraphics[width=0.115\linewidth]{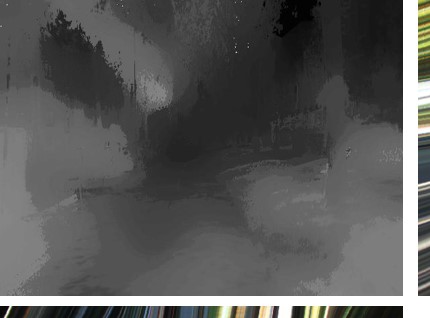}}~%
		
		\vspace{-7.0mm}
		\subfloat[]{\raisebox{5.5mm}{\rotatebox[origin=t]{90}{\cite{ivan2019synthesizing}}}}~%
		\subfloat[]{\includegraphics[width=0.115\linewidth]{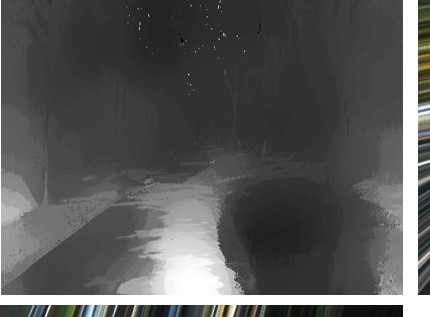}}~%
		\subfloat[]{\includegraphics[width=0.115\linewidth]{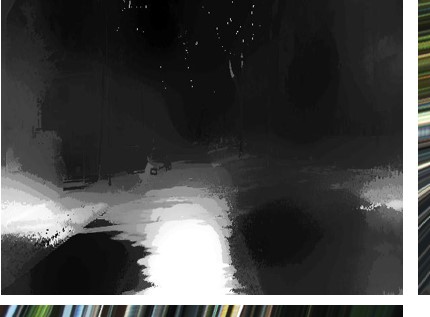}}~%
		\subfloat[]{\includegraphics[width=0.115\linewidth]{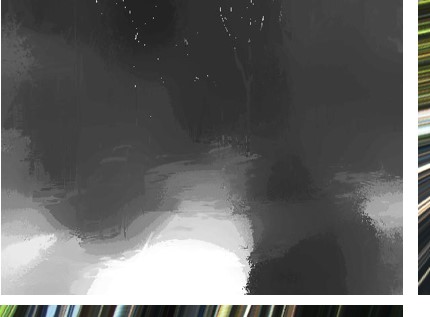}}~%
		\subfloat[]{\includegraphics[width=0.115\linewidth]{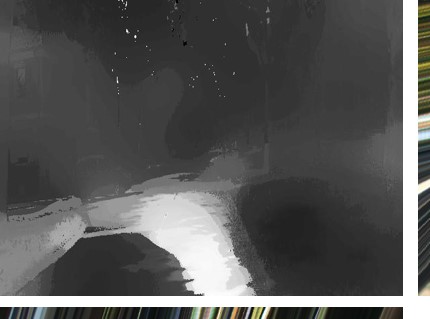}}~%
		\subfloat[]{\includegraphics[width=0.115\linewidth]{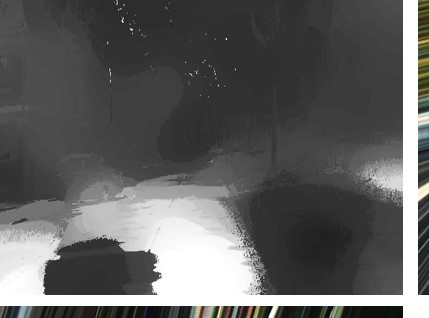}}~%
		\subfloat[]{\includegraphics[width=0.115\linewidth]{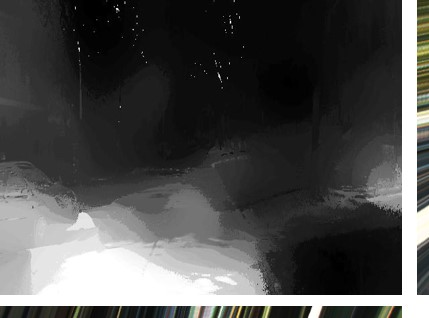}}~%
		\subfloat[]{\includegraphics[width=0.115\linewidth]{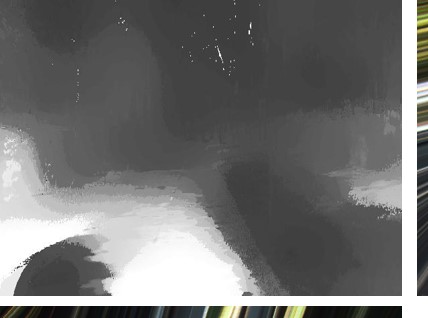}}~%
		\subfloat[]{\includegraphics[width=0.115\linewidth]{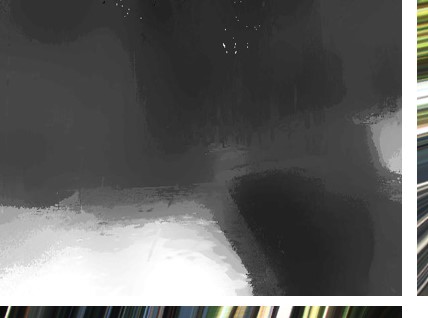}}~%
		
		\vspace{-7.0mm}
		\subfloat[]{\raisebox{5.0mm}{\rotatebox[origin=t]{90}{Proposed}}}~%
		\subfloat[]{\includegraphics[width=0.115\linewidth]{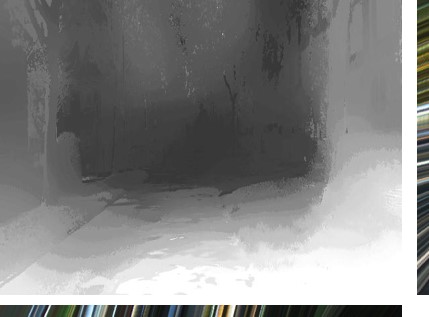}}~%
		\subfloat[]{\includegraphics[width=0.115\linewidth]{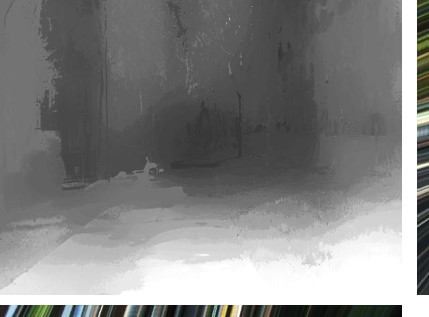}}~%
		\subfloat[]{\includegraphics[width=0.115\linewidth]{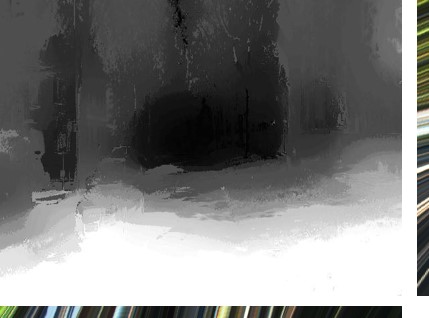}}~%
		\subfloat[]{\includegraphics[width=0.115\linewidth]{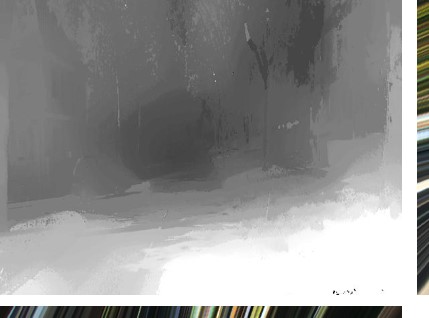}}~%
		\subfloat[]{\includegraphics[width=0.115\linewidth]{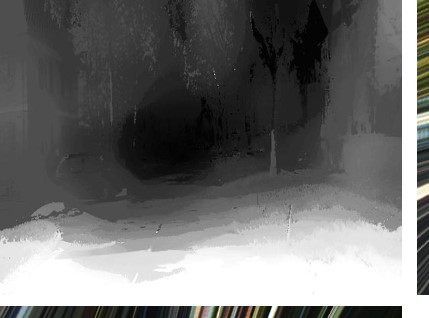}}~%
		\subfloat[]{\includegraphics[width=0.115\linewidth]{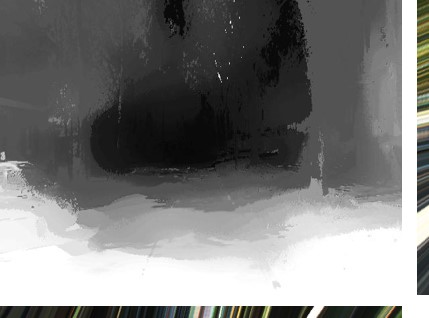}}~%
		\subfloat[]{\includegraphics[width=0.115\linewidth]{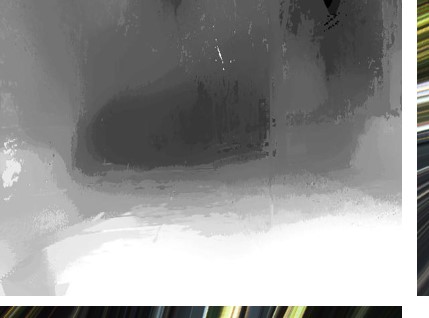}}~%
		\subfloat[]{\includegraphics[width=0.115\linewidth]{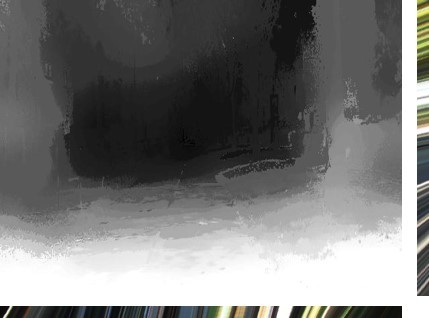}}~%
		\vspace{-8.0mm}
	\end{center}
	\caption{Qualitative comparison on the KITTI dataset. We show the eight continuous input video frames, their estimated depth after being synthesized as a light field frame using each method, and EPIs.}
	\label{fig:qualitative}
	\vspace{-4.0mm}
\end{figure*}

\begin{figure*}[t]
	\begin{center}
		{\includegraphics[width=0.25\linewidth]{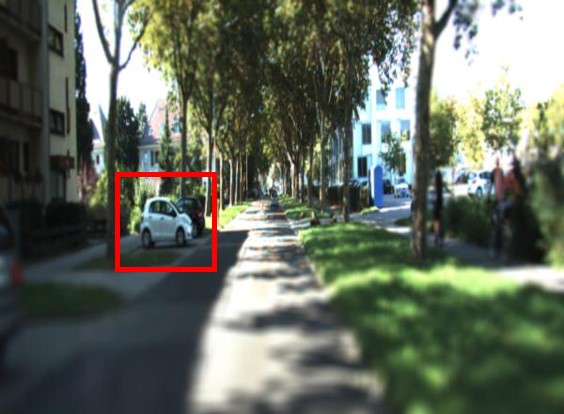}}~%
		{\includegraphics[width=0.25\linewidth]{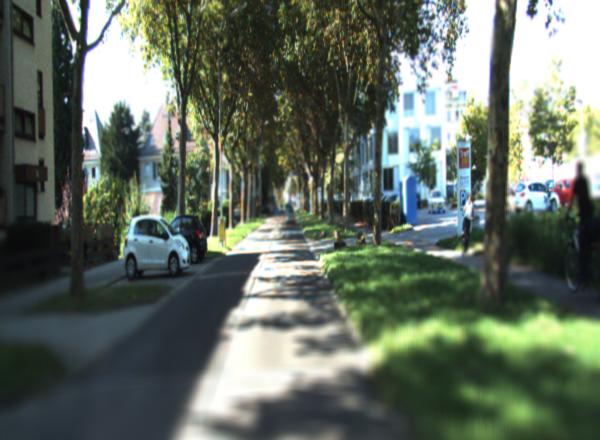}}~
		{\includegraphics[width=0.25\linewidth]{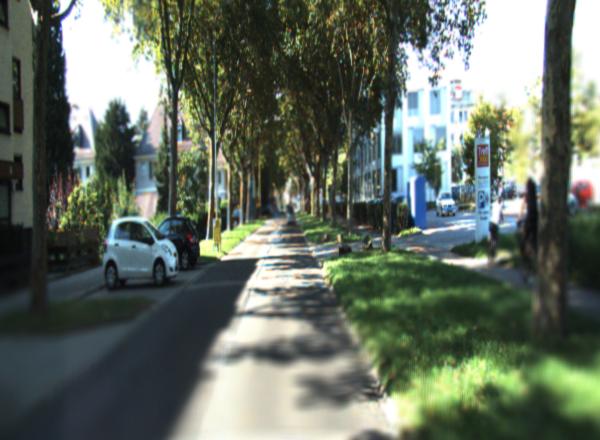}}~%
		{\includegraphics[width=0.25\linewidth]{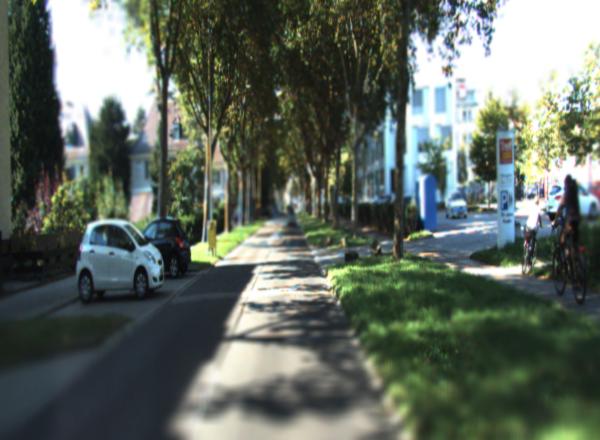}}~%
	\end{center}
	\vspace{-4.0mm}
	\caption{Video refocusing application. We show the four continuous frames focused on the white car on the left of the scene.}
	\label{fig:tracking}
	\vspace{-3.0mm}
\end{figure*}

We perform the qualitative evaluation of the proposed method on the KITTI dataset.
For comparison, after synthesizing the light field video from the input video, we estimate the depth using the synthesized light field video frames to evaluate the quality of synthesized light field video frame.
We use CAE~\cite{park2017robust}, which is the traditional light field depth estimation method.
Figure~\ref{fig:qualitative} shows the qualitative evaluation on the KITTI dataset.
As shown in Figure~\ref{fig:qualitative}, the results from \cite{srinivasan2017learning} show the insufficient degree of EPI slope and fail to estimate the accurate depth from the synthesized light field frame for some regions of the scene.
\cite{ivan2019synthesizing} failed to synthesize the light field given that it failed to estimate the accurate appearance flow for some regions of the scene.
In addition, \cite{srinivasan2017learning} and \cite{ivan2019synthesizing} synthesized temporally inconsistent light field video given that the estimated depth of the light field frames are not temporally consistent.
Note that, although it is trained with synthetic dataset, the proposed method synthesizes temporally consistent light field compared with other methods.
Moreover, we show the refocusing effect of the synthesized light field video in Figure~\ref{fig:tracking}.
\begin{table}[t]
	\centering
	\begin{tabular}{cclllllccllcllccllcllc}
		\hline
		& \multicolumn{6}{c}{Dataset}    &  & \multicolumn{6}{c}{Synthetic \#1}                                       &  & \multicolumn{6}{c}{Synthetic \#2}                                       &  \\ \hline
		& \multicolumn{6}{c}{Metric}     &  & \multicolumn{3}{c}{PSNR}           & \multicolumn{3}{c}{SSIM}           &  & \multicolumn{3}{c}{PSNR}           & \multicolumn{3}{c}{SSIM}           &  \\ \hline
		& \multicolumn{6}{c}{\cite{srinivasan2017learning}} &  & \multicolumn{3}{c}{22.56}          & \multicolumn{3}{c}{0.696}          &  & \multicolumn{3}{c}{24.99}          & \multicolumn{3}{c}{0.735}          &  \\
		& \multicolumn{6}{c}{\cite{ivan2019synthesizing}}       &  & \multicolumn{3}{c}{23.52}          & \multicolumn{3}{c}{0.708}          &  & \multicolumn{3}{c}{26.76}          & \multicolumn{3}{c}{0.804}          &  \\
		& \multicolumn{6}{c}{Proposed}   &  & \multicolumn{3}{c}{\textbf{23.77}} & \multicolumn{3}{c}{\textbf{0.732}} &  & \multicolumn{3}{c}{\textbf{27.11}} & \multicolumn{3}{c}{\textbf{0.831}} &  \\ \hline
	\end{tabular}
	\vspace{-2.0mm}
	\caption{Average PSNR (in dB) and SSIM for two synthetic test sets. Each test set consists of 162 frames.}
	\label{tab:quantitative}
	\vspace{-2.0mm}
\end{table}
\subsection{Quantitative Evaluation}\label{sec:4-3}
To evaluate the proposed method quantitatively, we use PSNR and SSIM~\cite{wang2004image}.
The average PSNR and SSIM for each test set are listed in Table~\ref{tab:quantitative}.
We use synthetic test sets \#1 and \#2 to perform a quantitative comparison between the proposed method and the existing state-of-the-art methods.
The test set \#1 has a similar appearance to the first row of Figure~\ref{fig:dataset}, and test set \#2 is similar to the second and third rows of Figure~\ref{fig:dataset}.
The test set \#1 is more challenging because it consists of specular reflection and high frequency components.
As shown in Table~\ref{tab:quantitative}, for the synthetic test set \#1, the proposed method outperforms the existing state-of-the-art methods by 1.2 and 0.2 dB in terms of PSNR and 0.03 and 0.02 in terms of SSIM.
For the synthetic test set \#2, the proposed method outperforms the previous methods by 2 and 0.3 dB in PSNR and 0.09 and 0.02 in SSIM.

\paragraph{Temporal Stability Evaluation}\label{sec:4-3-1}
We use Eq.~(\ref{eq:4.1}) to show that the synthesized light field video is temporally consistent.
Eq.~(\ref{eq:4.1}) warps an image at time ($t$) for each SAI to time ($t-1$) and calculates a warping error within a valid mask.
We perform this computation on all SAIs except for the center view because it is the same as the input frame.
Table \ref{tab:temporal error} shows the numerical values obtained by performing this process for the entire video and then averaging the warping error over the entire video.
As shown in Table \ref{tab:temporal error}, the proposed method can synthesize a light field video that is more stable in the temporal domain in comparison with existing methods.
\begin{table}[t]
	\centering
	\resizebox{\linewidth}{!}{
		\begin{tabular}{ccccccc}
			\hline
			& Dataset                                        &  & Synthetic \#1 &  & Synthetic \#2 &  \\ \hline
			& \cite{srinivasan2017learning} &  & 140.38        &  & 643.02        &  \\
			& \cite{ivan2019synthesizing}   &  & 139.45        &  & 642.70        &  \\
			& Proposed                                       &  & \textbf{138.72}        &  & \textbf{635.57}        &  \\ \hline
	\end{tabular}}
	\vspace{-2.0mm}
	\caption{Temporal stability evaluation for each test set.}
	\label{tab:temporal error}
	\vspace{-2.0mm}
\end{table}
\subsection{Ablation study}\label{sec:4-5}
To determine the effect of each loss function, we evaluate the proposed network by excluding each loss function $\ell_{temp}$, $\ell_{percep}$, and $\ell_{occ}$ and train the proposed network.
The ablation study for the loss functions~$\ell_{global}$ and $\ell_{local}$ which are proposed by \cite{ivan2019synthesizing} is not performed.
Table~\ref{tab:ablation} shows the quantitative evaluation when excluding each loss function one by one.
\begin{table}[t]
	\centering
	\resizebox{\linewidth}{!}{
		\begin{tabular}{cclllllccllcllccllcllc}
			\hline
			& \multicolumn{6}{c}{Dataset}  &  & \multicolumn{6}{c}{Synthetic \#1}                                       &  & \multicolumn{6}{c}{Synthetic \#2}                                       &  \\ \hline
			& \multicolumn{6}{c}{Metric}   &  & \multicolumn{3}{c}{PSNR}           & \multicolumn{3}{c}{SSIM}           &  & \multicolumn{3}{c}{PSNR}           & \multicolumn{3}{c}{SSIM}           &  \\ \hline
			& \multicolumn{6}{c}{without $\ell_{temp}$}   &  & \multicolumn{3}{c}{23.58}          & \multicolumn{3}{c}{0.688}          &  & \multicolumn{3}{c}{22.82}          & \multicolumn{3}{c}{0.646}          &  \\
			& \multicolumn{6}{c}{without $\ell_{percep}$}    &  & \multicolumn{3}{c}{21.47}          & \multicolumn{3}{c}{0.644}          &  & \multicolumn{3}{c}{24.21}          & \multicolumn{3}{c}{0.697}          &  \\
			& \multicolumn{6}{c}{without $\ell_{occ}$}    &  & \multicolumn{3}{c}{21.62}          & \multicolumn{3}{c}{0.632}          &  & \multicolumn{3}{c}{21.77}          & \multicolumn{3}{c}{0.602}          &  \\
			& \multicolumn{6}{c}{without $corr$}   &  & \multicolumn{3}{c}{23.45}          & \multicolumn{3}{c}{0.717}          &  & \multicolumn{3}{c}{24.52}          & \multicolumn{3}{c}{0.723}          &  \\
			& \multicolumn{6}{c}{Proposed} &  & \multicolumn{3}{c}{\textbf{23.77}} & \multicolumn{3}{c}{\textbf{0.732}} &  & \multicolumn{3}{c}{\textbf{27.11}} & \multicolumn{3}{c}{\textbf{0.831}} &  \\ \hline
	\end{tabular}}
	\vspace{-2.0mm}
	\caption{Quantitative evaluation of excluding each loss function's effect on the networks.}
	\label{tab:ablation}
	\vspace{-2.0mm}
\end{table}

\vspace*{-0.3cm}
\paragraph{Temporal Loss}
For the test set \#1, which has many non-Lambertian elements, when the temporal loss function $\ell_{temp}$ is excluded, the degradation is approximately 0.2 dB in PSNR and 0.04 in SSIM, respectively.
By contrast, for the dataset \#2 with a relatively simpler structure compared with dataset \#1, the difference in PSNR and SSIM is approximately 4 dB and 0.19, respectively.
This result indicates that the temporal consistency loss provides a considerable boost to the network performance for the simple scenes.

\vspace*{-0.3cm}
\paragraph{Perceptual Loss}
Contrary to $\ell_{temp}$, for the dataset with multiple non-Lambertian elements, the effect of excluding perceptual loss is approximately 2 dB in PSNR and 0.9 in SSIM, respectively.
For dataset \# 2, the effect of perceptual loss is relatively smaller than that of temporal loss; the effect is approximately 3 dB in PSNR and 0.13 in SSIM.
The perceptual loss works effectively on a scene that has many non-Lambertian elements rather than temporal loss.

\vspace*{-0.3cm}
\paragraph{Occlusion Loss}
In the case of $\ell_{occ}$, which indicates the influence of \textit{Occlusion Network} on the network, the effect is the greatest for datasets \# 1 and \#2, except for the PSNR value of the dataset \#1.
Excluding the network, the result shows that PSNR and SSIM values decrease to approximately 2 dB and 0.1 for dataset \#1 and approximately 5 dB and 0.2 for dataset \#2.

\vspace*{-0.3cm}
\paragraph{Correlation Layer}
We also train our network without correlation layer.
The effect of excluding the correlation layer is the least among the others.
The PSNR and SSIM values decrease to approximately 0.2 dB and 0.01 for the dataset \#1 and approximately 2.5 dB and 0.11 for dataset \#2.
It indicates that the correlation information from both frames works similar to temporal consistency loss.
Even though the effect is the least, excluding only a single correlation layer affects the performance significantly for simple scenes, such as test set \#2.

\section{Conclusion}
In this paper, we proposed a method for synthesizing 5D light field video with 9$\times$9 SAIs that is temporally consistent from a monocular video using synthetic light field datasets.
The proposed method considered the correlation between adjacent frames of the input monocular video and estimated the appearance flow, forward and backward optical flow using the extracted features.
The initial light field video was synthesized using the estimated appearance flow and provided temporal consistency to synthesized video using the estimated optical flow obtained using the mean image of the synthesized light field image.
In addition, a binary mask was formed using a variance image of the initial light field video frame.
It was used to improve the quality of occlusion and edge regions of the initial light field video frame using the proposed \textit{Occlusion Network}\.
The experimental results showed that the method was superior to the existing state-of-the-art methods quantitatively and qualitatively.
We analyzed the effects of each loss function through an ablation study and proposed an effective loss function were many non-Lambertian elements exist in the scene and when non.
The experimental results showed that the network trained using synthetic light field datasets can be generalized effectively for the datasets comprising actual scenes in addition to 3D graphic scenes.
We hope that our dataset and experimental results motivate researchers to solve the light field video synthesis problem.

\section*{Acknowledgement}
This work is supported by Samsung Research Funding Center of Samsung Electronics under Project Number SRFC-IT1702-06.
{\small
	\bibliographystyle{ieee_fullname}
	\bibliography{egbib}
}
\end{document}